\documentclass[lineno]{biometrika}

\usepackage{amsmath}

\usepackage{times}
\usepackage{bm}
\usepackage{natbib}

\usepackage{amssymb}

\usepackage[plain,noend]{algorithm2e}

\makeatletter
\renewcommand{\algocf@captiontext}[2]{#1\algocf@typo. \AlCapFnt{}#2} 
\def\@algocf@capt@plain{top}
\renewcommand{\algocf@makecaption}[2]{%
  \addtolength{\hsize}{\algomargin}%
  \sbox\@tempboxa{\algocf@captiontext{#1}{#2}}%
  \ifdim\wd\@tempboxa >\hsize
    \hskip .5\algomargin%
    \parbox[t]{\hsize}{\algocf@captiontext{#1}{#2}}
  \else%
    \global\@minipagefalse%
    \hbox to\hsize{\box\@tempboxa}
  \fi%
  \addtolength{\hsize}{-\algomargin}%
}
\makeatother

\begin{document}



\markboth{N. Binkiewicz, J. T. Vogelstein \and K. Rohe}{Covariate-assisted spectral clustering}

\title{Covariate-assisted spectral clustering}

\author{N. BINKIEWICZ}
\affil{Department of Statistics, University of Wisconsin, 1300 University Ave., Madison 53706, U.S.A. \email{norbertbin@gmail.com}}

\author{J. T. VOGELSTEIN}
\affil{Department of Biomedical Engineering and Institute for Computational Medicine, Johns Hopkins University, 720 Rutland Ave., Baltimore 21205, U.S.A. \email{jovo@jhu.edu}}

\author{\and K. ROHE}
\affil{Department of Statistics, University of Wisconsin, 1300 University Ave., Madison 53706, U.S.A. \email{karlrohe@stat.wisc.edu}}

\maketitle

\begin{abstract}
Biological and social systems consist of myriad interacting units. The interactions can be represented in the form of a graph or network. Measurements of these graphs can reveal the underlying structure of these interactions, which provides insight into the systems that generated the graphs. Moreover, in applications such as connectomics, social networks, and genomics, graph data are accompanied by contextualizing measures on each node. We utilize these node covariates to help uncover latent communities in a graph, using a modification of spectral clustering. Statistical guarantees are provided under a joint mixture model that we call the node-contextualized stochastic blockmodel, including a bound on the mis-clustering rate. The bound is used to derive conditions for achieving perfect clustering. For most simulated cases, covariate-assisted spectral clustering yields results superior to regularized spectral clustering without node covariates and to an adaptation of canonical correlation analysis. We apply our clustering method to large brain graphs derived from diffusion MRI data, using the node locations or neurological region membership as covariates. In both cases, covariate-assisted spectral clustering yields clusters that are easier to interpret neurologically.
\end{abstract}

\begin{keywords}
Brain graph; Laplacian; Network; Node attribute; Stochastic blockmodel.
\end{keywords}

\section{INTRODUCTION}\label{intro}

Modern experimental techniques in areas such as genomics and brain imaging generate vast amounts of structured data, which contain information about the relationships of genes or brain regions. Studying these relationships is essential for solving challenging scientific problems, but few computationally feasible statistical techniques incorporate both the structure and diversity of these data. 

A common approach to understanding the behavior of a complex biological or social system is to first discover blocks of highly interconnected units, also known as communities or clusters, that serve or contribute to a common function. These might be genes that are involved in a common pathway or areas in the brain with a common neurological function. Typically, we only observe the pairwise relationships between the units, which can be represented by a graph or network. Analyzing networks has become an important part of the social and biological sciences. Examples of such networks include gene regulatory networks, friendship networks, and brain graphs. If we can discover the underlying block structure of such graphs, we can gain insight from the common characteristics or functions of the units within a block.  

Existing research has extensively studied the algorithmic and theoretical aspects of finding node clusters within a graph, by Bayesian, maximum likelihood, and spectral approaches. Unlike model-based methods, spectral clustering is a relaxation of a cost minimization problem and has shown to be effective in various settings \citep{ng2001spectral,von2007tutorial}. Modifications of spectral clustering, such as regularized spectral clustering, are accurate even for sparse networks \citep{chaudhuri2012spectral,amini2013pseudo, qin2013regularized}. On the other hand, certain Bayesian methods offer additional flexibility in how nodes are assigned to blocks, allowing for a single node to belong to multiple blocks or a mixture of blocks \citep{nowicki2001estimation,airoldi2008mixed}. Maximum likelihood approaches can enhance interpretabilty by embedding nodes in a latent social space and providing methods for quantifying statistical uncertainty \citep{hoff2002latent,handcock2007model, amini2013pseudo}. For large graphs, spectral clustering is one of very few computationally feasible methods that has an algorithmic guarantee for finding the globally optimal partition.

The structured data generated by modern technologies often contain additional measurements that can be represented as graph node attributes or covariates. For example, these could be personal profile information in a friendship network or the spatial location of a brain region in a brain graph. There are two potential advantages of utilizing node covariates in graph clustering. First, if the covariates and the graph have common latent structure, then the node covariates provide additional information to help estimate this structure. Even if the covariates and the graph do not share exactly the same structure, some similarity is sufficient for the covariates to assist in the discovery of the graph structure. Second, by using node covariates in the clustering procedure, we enhance the relative homogeneity of covariates within a cluster and filter out partitions that fail to align with the important covariates. This allows for easy contextualization of the clusters in terms of the member nodes' covariates, providing a natural way to interpret the clusters.

Methods that utilize both node covariates and the graph to cluster the nodes have previously been introduced, but many of them rely on ad hoc or heuristic approaches and none provide theoretical guarantees for statistical estimation. Most existing methods can be broadly classified into Bayesian approaches, spectral techniques, and heuristic algorithms. Many Bayesian models focus on categorical node covariates and are often computationally expensive \citep{chang2010hierarchical, balasubramanyan2011block}. A recent Bayesian model proposed by \citet{yang2014community} can discover multi-block membership of nodes with binary node covariates. This method has linear update time in the network size, but does not guarantee linear-time convergence. Heuristic algorithms use various approaches, including embedding the network in a vector space, at which point more traditional methods can be applied to the vector data \citep{gibert2012graph} or using the covariates to augment the graph and applying other graph clustering methods that tune the relative weights of node-to-node and node-to-covariate edges \citep{zhou2009graph}. A commonly-used spectral approach to incorporate node covariates directly alters the edge weights based on the similarity of the corresponding nodes' covariates, and uses traditional spectral clustering on the weighted graph \citep{neville2003clustering,gunnemann2013spectral}. 

This work introduces a spectral approach that performs well for assortative graphs and another that does not require this restriction. We give a standard definition of an assortative graph here and later define it in the context of a stochastic blockmodel.

\begin{definition}\label{assortGeneral}
(Assortative graph) A graph is assortative if nodes within the same cluster are more likely to share an edge than nodes in two different clusters. 
\end{definition}

Assortative covariate-assisted spectral clustering adds the covariance matrix of the node covariates to the regularized graph Laplacian, boosting the signal in the top eigenvectors of the sum, which is then used for spectral clustering. This works well for assortative graphs, but performs poorly otherwise. Covariate-assisted spectral clustering, which uses the square of the regularized graph Laplacian, is presented as a more general method that performs well for assortative and non-assortative graphs. A tuning parameter is employed by both methods to adjust the relative weight of the covariates and the graph; \S \ref{oProc} proposes a way to choose this tuning parameter. Research on dynamic networks using latent space models has yielded an analogous form for updating latent coordinates based on a distance matrix and the latent coordinates from the previous time step \citep{sarkar2006dynamic}. A similar framework can also be used to cluster multiple graphs \citep{eynard2012multimodal}. 

Variants of our methods previously introduced were derived by first considering an optimization problem to minimize the weighted sum of the k-means and graph cut objective functions and then solving a spectral relaxation of the original problem. \citet{wang2009integrated} decided against using an additive method similar to covariate-assisted spectral clustering because setting the method's tuning parameter is a non-convex problem. They chose to investigate a method that uses the product of the generalized inverse of the graph Laplacian and the covariate matrix instead. \citet{shiga2007spectral} recognized the advantage of having a tuning parameter to balance the contribution of the graph and the covariates, but did not use the stochastic blockmodel to study their method. The full utility and flexibility of these types of approaches have not yet been presented, and neither paper derives any statistical results about their performance. Furthermore, they do not consider the performance of these methods on non-assortative graphs. In contrast, we were initially motivated to develop covariate-assisted spectral clustering by its interpretation and propensity for theoretical analysis.

Very few of the clustering methods that employ both node covariates and the graph offer any theoretical results, and, to our knowledge, this paper gives the first statistical guarantee for these types of approaches. We define the node-contextualized stochastic blockmodel, which combines the stochastic blockmodel with a block mixture model for node covariates. Under this model, a bound on the mis-clustering rate of covariate-assisted spectral clustering is established in \S \ref{mcBoundSection}. The behavior of the bound is studied for a fixed and an increasing number of covariates as a function of the number of nodes, and conditions for perfect clustering are derived. A general lower bound is also derived, demonstrating the conditions under which an algorithm using both the node covariates and the graph can give more accurate clusters than any algorithm using only the node covariates or the graph. 

For comparison, an alternative method based on an adaptation of classical canonical correlation analysis is introduced \citep{hotelling1936relations}, which uses the product of the regularized graph Laplacian and the covariate matrix as the input to the spectral clustering algorithm. Simulations indicate that canonical correlation performs worse than covariate-assisted spectral clustering under the node-contextualized stochastic blockmodel with Bernoulli covariates. However, canonical correlation analysis clustering is computationally faster than our clustering method and requires no tuning. In contrast, covariate-assisted spectral clustering depends on a single tuning parameter, which interpolates between spectral clustering with only the graph and only the covariates. This parameter can be set without prior knowledge by using an objective function such as the within-cluster sum of squares. Some results for determining what range of tuning parameter values should be considered are provided in the description of the optimization procedure in \S \ref{oProc}. Alternatively, the tuning parameter can be set using prior knowledge or to ensure the clusters achieve some desired quality, such as spatial cohesion. As an illustrative example, \S \ref{dataSection} studies diffusion magnetic resonance imaging derived brain graphs using two different sets of node covariates. The first analysis uses spatial location. This produces clusters that are more spatially coherent than those obtained using regularized spectral clustering alone, making them easier to interpret. The second analysis uses neurological region membership, which yields partitions that closely align with neurological regions while allowing for patient-wise variability based on brain graph connectivity. 

\section{METHODOLOGY}

\subsection{Notation}

Let $G(E,V)$ be a graph, where $V$ is the set of vertices or nodes and $E$ is the set of edges, which represent relationships between the nodes. Let $N$ be the number of nodes. Index the nodes in $V = \{1, \dots , N \}$; then $E$ contains a pair $(i, j)$ if there is an edge between nodes $i$ and $j$. A graph's edge set can be represented as the adjacency matrix $A \in \{0, 1\}^{N \times N}$, where $A_{ij} = A_{ji} = 1$ if $(i, j) \in E$ and $A_{ij} = A_{ji} = 0$ otherwise. We restrict ourselves to studying undirected and unweighted graphs, although with small modifications most of our results apply to directed and weighted graphs as well. 

Define the regularized graph Laplacian as 
\begin{align*}
	L_{\tau}  = D_{\tau}^{-1/2} A D_{\tau}^{-1/2} ,
\end{align*}
where $D_{\tau} = D + \tau I$ and $D$ is a diagonal matrix with $D_{ii} = \sum_{j=1}^N A_{ij}$. The regularization parameter $\tau$ is treated as a constant, and is included to improve spectral clustering performance on sparse graphs \citep{chaudhuri2012spectral}. Throughout, we shall set $\tau = N^{-1} \sum_{i=1}^N D_{ii}$, i.e., the average node degree \citep{qin2013regularized}.
 
For the graph $G(E,V)$, let each node in the set $V$ have an associated bounded covariate vector $X_i \in [-J,J]^{R}$, and let $X \in [-J,J]^{N \times R}$ be the covariate matrix where each row corresponds to a node covariate vector. Let $\parallel \cdot \parallel$ denote the spectral norm and $\parallel \cdot \parallel_F$ denote the Frobenius norm. Let $I(\cdot)$ denote the indicator function. For a sequence $\{ a_N\}$ and $\{ b_N \}$, $a_N = \Theta(b_N)$ if and only if $b_N = O(a_N)$ and $a_N = O(b_N)$. 

\subsection{Spectral clustering for a graph with node covariates}

The spectral clustering algorithm has been employed to cluster graph nodes using various functions of the adjacency matrix. For instance, applying the algorithm to $L_{\tau} $ corresponds to regularized spectral clustering, where the value of the regularization parameter is set prior to running the algorithm. All of the methods we consider will employ this algorithm, but will use a different input matrix such as $L_{\tau} $, $\tilde{L} $, or $L^{\rm CCA}$ as defined later.

\begin{algo} \label{al1}
Spectral clustering. \\
Given input matrix $W$ and number of clusters $K$: \\
\textit{Step} 1. Find eigenvectors $U_1, ..., U_K \in \mathbb{R}^N$ corresponding to the $K$ largest eigenvalues of $W$.\\
\textit{Step} 2. Use the eigenvectors as columns to form the matrix $U = [ U_1, ..., U_K ] \in \mathbb{R}^{N \times K}$. \\
\textit{Step} 3. Form the matrix $U^*$ by normalizing each of $U$'s rows to have unit length. \\
\textit{Step} 4. Run $k$-means clustering with $K$ clusters treating each row of $U^*$ as a point in $\mathbb{R}^K$. \\
\textit{Step} 5. If the $i$th row of $U^*$ falls in the $k$th cluster, assign node $i$ to cluster $k$.
\end{algo}

Step 4 of the spectral clustering algorithm uses $k$-means clustering, which is sensitive to initialization. In order to reduce this sensitivity, we use multiple random initializations.
To take advantage of available graph and node covariate data in graph clustering, it is necessary to employ methods that incorporate both of these data types. As discussed in \S \ref{intro}, spectral clustering has many advantages over other graph clustering methods. Hence, we propose three approaches that use the spectral clustering framework and utilize both the graph structure and the node covariates. 

Assortative covariate-assisted spectral clustering uses the leading eigenvectors of
\begin{align*}
	\bar{L} (\alpha) = L_{\tau}  + \alpha X X^T,
\end{align*}
where $\alpha \in [0, \infty)$ is a tuning parameter. When using $\{0,1\}$-Bernoulli covariates, the covariate term can be interpreted as adding to each element $(i,j)$ a value proportional to the number of covariates equal to one for both $i$ and $j$. In practice, the covariate matrix $X$ should be parameterized as in linear regression; specifically, categorical covariates should be re-expressed with dummy variables. For continuous covariates, it can be beneficial to center and scale the columns of $X$ before performing the analysis. The scaling helps satisfy condition $(i)$ in Lemma \ref{blockLemma}, which ensures the top $K$ eigenvectors of $\bar{L} (\alpha)$ contain block information. As demonstrated in the simulations in \S \ref{simSection}, this method is robust and has good performance for assortative graphs, but does not perform well for non-assortative graphs. 

Covariate-assisted spectral clustering uses the leading eigenvectors of
\begin{align*}
	\tilde{L} (\alpha) = L_{\tau} L_{\tau}  + \alpha X X^T.
\end{align*}
This approach performs well for non-assortative graphs and nearly as well as our assortative clustering method for assortative graphs. When there is little chance of confusion, $\tilde{L}$ will be used for notational convenience.

To run covariate-assisted spectral clustering on the large graphs, such as the brain graphs in \S \ref{dataSection}, the top $K$ eigenvectors of $\tilde{L} $ are computed using the implicitly restarted Lanczos bidiagonalization algorithm \citep{baglama2006restarted}. At each iteration, the algorithm only needs to compute the product $\tilde{L}  v$, where $v$ is an arbitrary vector. For computational efficiency, the product is calculated as $L_{\tau} (L_{\tau}  v) + \alpha X (X^T v)$. This takes advantage of the sparsity of $L_{\tau} $ and the low-rank structure of $XX^T$. Ignoring log terms and any special structure in $X$, it takes $O\{(|E|+NR)K\}$ operations to compute the required top $K$ eigenvectors of $\tilde{L} $, where $R$ is the number of columns in $X$. The graph clusters are obtained by iteratively employing the spectral clustering algorithm on $\tilde{L} (\alpha)$ while varying the tuning parameter $\alpha$ until an optimal value is obtained. The details of this procedure are described in the next section.

As an alternative, we propose a modification of classical canonical correlation analysis \citep{hotelling1936relations} whose similarity matrix is the product of the regularized graph Laplacian and the covariate matrix,
\begin{align*}
	L^{\rm CCA} = L_{\tau}  X.
\end{align*} 
The spectral clustering algorithm is employed on $L^{\rm CCA}$ to obtain node clusters when the number of covariates, $R$, is greater than or equal to the number of clusters, $K$. This approach inherently provides a dimensionality reduction in the common case where the number of covariates is much less than the number of nodes. If $R \ll N^{-1} \sum_i D_{ii}$, then spectral clustering with $L^{\rm CCA}$ has a faster running time than covariate-assisted spectral clustering.

\subsection{Setting the tuning parameter}\label{oProc}
In order to preform spectral clustering with $\tilde{L} (\alpha)$, it is necessary to determine a specific value for the tuning parameter, $\alpha$. The tuning procedure presented here presumes that both the graph and the covariates contain some block information, as demonstrated by the simulations in \S \ref{simSection}. In practice, an initial test can be used to determine if the graph and the covariates contain common block information, and such a test will be presented in future work. The tuning parameter should be chosen to achieve a balance between $L_{\tau} $ and $X$ such that the information in both is captured in the leading eigenspace of $\tilde{L} $. For large values of $\alpha$, the leading eigenspace of $\tilde{L} $ is approximately the leading eigenspace of $XX^T$. For small values of $\alpha$, the leading eigenspace of $\tilde{L} $ is approximately the leading eigenspace of $L_{\tau}$. A good initial choice of $\alpha$ is that which makes the leading eignevalues of $L_{\tau}L_{\tau}$ and $\alpha XX^T$ equal, namely $\alpha_0 = \lambda_1(L_{\tau} L_{\tau})/\lambda_1(XX^T)$.

There is a finite range of $\alpha$ where the leading eigenspace of $\tilde{L} (\alpha)$ is not a continuous function of $\alpha$; outside this range, the leading eigenspace is always continuous in $\alpha$. In simulations, the clustering results are exceedingly stable in the continuous range of $\alpha$. Hence, only the values of $\alpha$ inside a finite interval need to be considered. This section gives an interval $\alpha \in [\alpha_{\rm min}, \alpha_{\rm max}]$ that is computed with only the eigenvalues of $L_\tau L_\tau$ and $XX^T$. Within this interval, $\alpha$ is chosen to minimize an objective function. Empirical results demonstrating these properties are given in the Supplementary Material.

Let $\lambda_i(M)$ be the $i$th eigenvalue of matrix $M$. To find the initial range $[\alpha_{\rm min}, \alpha_{\rm max}]$, define a static vector $v \in \mathbb{R}^N$ as a vector that satisfies either condition (a) or (b) below. For $\epsilon \geq 0$,
\begin{align*}
&\text{(a) } v^T L_{\tau} L_{\tau} v \geq \lambda_K(L_{\tau} L_{\tau}),\text{ } v^TXX^T v \leq \epsilon, \\
&\text{(b) } v^TXX^T v \geq \lambda_K(XX^T),\text{ } v^T L_{\tau} L_{\tau} v\leq \epsilon.
\end{align*}
These are vectors for which $XX^T$ and $L_{\tau} L_{\tau}$ are highly differentiated; perhaps there is a cluster in the graph that does not appear in the covariates, or vice versa. These static vectors produce discontinuities in the leading eigenspace of $\tilde{L}(\alpha)$.

For example, let $v_*$ be an eigenvector of $L_{\tau} L_{\tau}$ and a static vector of type (a), then as $\alpha$ changes, it will remain a slightly perturbed eigenvector of $\tilde{L}(\alpha)$.  When $v^T_* \tilde{L}(\alpha_*) v_*$ is close to $\lambda_K\{\tilde{L}(\alpha_*)\}$, then in some neighborhood of $\alpha_*$, the slightly perturbed version of $v_*$ will transition into the leading eigenspace of $\tilde{L}$.  This transition corresponds to a discontinuity in the leading eigenspace.

As shown in the Supplementary Material, the concept of static vectors with $\epsilon = 0$ can be used to find a limited range of $\alpha$ for possible discontinuities. The range of $\alpha$ values for which discontinuities can occur is $[\alpha_{\rm min}, \alpha_{\rm max}]$, where 
\begin{align*}
	&\alpha_{\rm min} = \frac{\lambda_{K}(L_{\tau} L_{\tau}) - \lambda_{K+1}(L_{\tau} L_{\tau})}{\lambda_1(X X^T)},\\
 	&\alpha_{\rm max} = \frac{\lambda_{1}(L_{\tau} L_{\tau})}{\lambda_{R}(XX^T) I(R \leq K) + \{\lambda_{K}(XX^T) - \lambda_{K+1}(X X^T)\} I(R > K)}.
\end{align*}

The tuning parameter $\alpha \in [\alpha_{\rm min}, \alpha_{\rm max}]$ is chosen to be the value which minimizes the k-means objective function, the within cluster sum of squares, 
\begin{align*}
	\Phi(\alpha) = \sum_{i=1}^K \sum_{u_j \in F_i} ||u_j(\alpha) - C_i(\alpha) ||^2 ,
\end{align*}
where $u_j$ is the $j$th row of $U$, $C_i$ is the centroid of the $i$th cluster from k-means clustering, and $F_i$ is the set of points in the $i$th cluster. Hence, the tuning parameter is $\alpha = \text{\rm argmin}_{\alpha \in [\alpha_{\rm min}, \alpha_{\rm max}]} \Phi(\alpha)$.

\section{THEORY}
\subsection{Node-contextualized stochastic blockmodel}
To illustrate what covariate-assisted spectral clustering estimates, this section proposes a statistical model for a network with node covariates and shows that covariate-assisted spectral clustering is a weakly consistent estimator of certain parameters in the proposed model. To derive statistical guarantees for covariate-assisted spectral clustering, we assume a joint mixture model for the the graph and the covariates. Under this model, each node belongs to one of $K$ blocks and each edge in the graph corresponds to an independent Bernoulli random variable. The probability of an edge between any two nodes depends only on the block membership of those nodes \citep{holland1983stochastic}. In addition, each node is associated with $R$ independent covariates with bounded support, where expectation depends only on the block membership and $R$ can grow with the number of nodes.

\begin{definition}\label{SBM}
(Node-contextualized stochastic blockmodel) Consider a set of nodes $\{1, ..., N\}$. Let $Z \in \{0,1\}^{N \times K}$ assign the $N$ nodes to one of the $K$ blocks, where $Z_{ij} = 1$ if node $i$ belongs to block $j$. Let $B \in [0,1]^{K \times K}$ be full rank and symmetric, where $B_{ij}$ is the probability of an edge between nodes in blocks $i$ and $j$. Conditional on $Z$, the elements of the adjacency matrix are independent Bernoulli random variables. The population adjacency matrix $\mathcal{A} = E(A \mid Z)$ fully identifies the distribution of $A$ and $\mathcal{A} = Z B Z^T$.

Let $X \in [-J,J]^{N \times R}$ be the covariate matrix and $M \in [-J,J]^{K \times R}$ be the covariate expectation matrix, where $M_{i,j}$ is the expectation of the $j$th covariate of a node in the $i$th block. Conditional on $Z$, the elements of $X$ are independent and the population covariate matrix, $\mathcal{X} = E(X \mid Z)$, is 
\begin{align}\label{covModel}
	\mathcal{X} = Z M.
\end{align}
\end{definition}

Under the node-contextualized stochastic blockmodel, covariate-assisted spectral clustering seeks to estimate the block membership matrix $Z$. In the next section, we show its consistency. If $B$ is assumed to be positive definite, the same results hold for assortative covariate-assisted spectral clustering up to a constant factor. These results motivate the definition of an assortative graph in the context of the node-contextualized stochastic blockmodel. 

\begin{definition}\label{assort}
(Assortative graph) A graph generated under the node-contextualized stochastic blockmodel is said to be assortative if the block probability matrix $B$ corresponding to the graph is positive definite. Otherwise, it is said to be non-assortative. 
\end{definition}

Many common networks are assortative, such as friendship networks or brain graphs. Dating networks are one example of a non-assortative network. Most relationships in a dating network are heterosexual, comprised of one male and one female. In a stochastic blockmodel, where the blocks are constructed by gender, $B$ will have small diagonal elements and large off-diagonal elements, producing more relationships between genders than within genders. Such a matrix is not positive definite. More generally, non-assortative stochastic blockmodels will tend to generate more edges between blocks and fewer edges within blocks. These non-assortative blocks appear in the spectrum of $L_\tau$ as large negative eigenvalues. By squaring the matrix $L_\tau$, the eigenvalues become large and positive, matching the positive eigenvalues in $XX^T$.

\subsection{Covariate-assisted spectral clustering is statistically consistent under the node-contextualized stochastic blockmodel}\label{mcBoundSection}
The proof of consistency for covariate-assisted spectral clustering under the node-contextualized stochastic blockmodel requires three results. Lemma \ref{blockLemma} expresses the eigendecomposition of the population version of the covariate-assisted Laplacian,
\begin{align*}
\tilde{\mathcal{L}}(\alpha) = (\mathcal{D} + \tau I)^{-1/2} \mathcal{A} (\mathcal{D} + \tau I)^{-1} \mathcal{A} (\mathcal{D} + \tau I)^{-1/2} + \alpha E(XX^T),
\end{align*}
in terms of $Z$. Theorem \ref{concentration} bounds the spectral norm of the difference between $\tilde{L}$ and $\tilde{\mathcal{L}}$. Then, the Davis--Kahan Theorem \citep{davis1970rotation} bounds the difference between the sample and population eigenvectors in Frobenius norm. Finally, Theorem \ref{mcRate} combines these results to establish a bound on the mis-clustering rate of covariate-assisted spectral clustering. The argument largely follows \cite{qin2013regularized}. The results provided here do not include the effects of Step 3 in Algorithm 1. The proofs are in the Supplementary Material.

\begin{lemma}\label{blockLemma}
(Equivalence of eigenvectors and block membership) 

Under the node-contextualized stochastic blockmodel, let $\mathcal{D}_B = \text{diag}(BZ^T\mathbf{1}_n + \tau)$, $\tilde{P}=Z^T Z$, and $\tilde{B} = \mathcal{D}_B^{-1/2} B Z^T (\mathcal{D} + \tau I)^{-1} Z B \mathcal{D}_B^{-1/2} + \alpha M M^T$. Let $\varkappa = \max_l |c_l-\bar{c}|$, where $c_l = \sum_i Var(X_{il}|Z_i=l)$ and $\bar{c} = \sum_l c_l/K$. Define $\mathcal{U} \in \mathbb{R}^{N \times K}$ with columns containing the top $K$ eigenvectors of $\tilde{\mathcal{L}}$. Assume $\text{(i) } \lambda_K(\tilde{B}\tilde{P}) > 2 \alpha \varkappa$, then there exists an orthogonal matrix $V \in \mathbb{R}^{K \times K}$, such that $\mathcal{U} = Z(ZZ^T)^{-1/2}V$. Furthermore,
$Z_i(ZZ^T)^{-1/2}V = Z_j(ZZ^T)^{-1/2}V$ if and only if $Z_i = Z_j$, where $Z_i$ is the $i$th row of the block membership matrix.
\end{lemma}

Under assumption $(i)$, the rows of the population eigenvectors are equal if and only if the corresponding nodes belong to the same block. This assumption requires the population eigengap to be greater than the maximum of the absolute difference between the sum of covariate variances within a block and the mean of the sums across all blocks. If all the covariates have equal variance in all blocks, the assumption is trivially true. Since $XX^T$ is effectively being used as a measure of similarity between nodes, if the covariate variances across blocks are unequal, the difference in scale makes the blocks more difficult to distinguish. This is evidenced by a reduction in the eigengap proportional to this difference. In practice, this condition is not restrictive since $X$ can be centered and normalized. To derive a bound on the mis-clustering rate, we will need a bound on the difference between the population eigenvectors and the sample eigenvectors. In order to establish this bound, the following theorem bounds the spectral norm of the difference between $\tilde{L}$ and $\tilde{\mathcal{L}}$. 

\begin{theorem}\label{concentration}
(Concentration inequality) Let $d=\min \mathcal{D}_{ii}$, $\mathcal{X}_{ik}^{(p)} = E(X_{ik}^p)$, $\varpi = 8 \alpha^2 \sum_k \left\{ \sum_i \mathcal{X}_{ik}^{(2)} \sum_l ( \mathcal{X}_{lk}^{(2)} - \mathcal{X}_{lk}^2) + \mathcal{X}_{ik}^{(4)} \right\}$, 
$\delta = 12(d+\tau)^{-1/2}  + \varpi^{1/2}$, $S = 3\alpha NJ^2$.
For any $\epsilon > 0$, if $\text{(ii) } d + \tau > 3 \log(8N/\epsilon)$ and $\text{(iii) } \varpi/S^2 > 3 \log(8N/\epsilon)$ then with probability at least $1 - \epsilon$,
\begin{align*}
 ||\tilde{L} - \tilde{\mathcal{L}} || \leq \delta \{3\log (8N/\epsilon)\}^{1/2}.
\end{align*}
\end{theorem}

Consider a node-contextualized stochastic blockmodel with two blocks, within block probabilities $p$, and between block probabilities $q$. Condition $(ii)$ holds when $p + q > \Theta\{\log(N)/N\}$ and condition $(iii)$ holds when $R \geq \Theta(\log N)$. Hence, condition $(ii)$ restricts the sparsity of the graph, while condition $(iii)$ requires that the number of covariates grows with the number of nodes. Now we use Theorem \ref{concentration} and the Davis--Kahan Theorem to bound the difference between the sample and population eigenvectors.

\begin{theorem}\label{evBound}
(Empirical and population eigenvector bound) Let $\lambda_K$ be the $K$th largest eigenvalue of $\tilde{L}$ and $\mathcal{O}$ be a rotation matrix. Let the columns of $U$ and $\mathcal{U}$ contain the top $K$ eigenvectors of $\tilde{L}$ and $\tilde{\mathcal{L}}$, respectively. Given assumptions (i) in Lemma \ref{blockLemma}, (ii) and (iii) in Theorem \ref{concentration}, and (iv) $\delta \{3 \log(8N/\epsilon)\}^{1/2}  \leq \lambda_K/2$, then with probability at least $1 - \epsilon$, 
\begin{align*}
\parallel U - \mathcal{U} \mathcal{O} \parallel_F \leq \frac{8 \delta \{3 K \log(8N/\epsilon)\}^{1/2}}{\lambda_K}.
\end{align*}
\end{theorem} 

The next theorem bounds the proportion of mis-clustered nodes. In order to define mis-clustering, recall that the spectral clustering algorithm uses k-means clustering to cluster the rows of $U$. Let $C_i$ and $\mathcal{C}_i$ be the cluster centroid of the $i$th node generated using k-means clustering on $U$ and $\mathcal{U}$, respectively. A node $i$ is correctly clustered if $C_i$ is closer to $\mathcal{C}_i$ than $\mathcal{C}_j$ for all $j$ such that $Z_j \neq Z_i$. In order to avoid identifiablity problems and since clustering only requires the estimation of the correct subspace, the formal definition is augmented with a rotation matrix $\mathcal{O}$. The following definition formalizes this intuition.

\begin{definition}
(Set of mis-clustered nodes)
Let $\mathcal{O}$ be a rotation matrix that minimizes $||U\mathcal{O}^T - \mathcal{U}||_F$. Define the set of mis-clustered nodes as
\begin{align*}
 \mathcal{M} = \{i: \text{ there exists } j \neq i \text{ such that } ||C_i\mathcal{O}^T - \mathcal{C}_i||_2 > ||C_i \mathcal{O}^T - \mathcal{C}_j||_2 \}.
\end{align*}
\end{definition}

Using the definition of mis-clustering and the result from Theorem \ref{evBound}, the next theorem bounds the mis-clustering rate, $| \mathcal{M} | / N$.

\begin{theorem}\label{mcRate}
(Mis-clustering rate bound) Let $P = \max_i (Z^TZ)_{ii}$ denote the size of the largest block. Under assumptions (i) in Lemma \ref{blockLemma}, (ii) and (iii) in Theorem \ref{concentration}, and (iv) in Theorem \ref{evBound}, with probability at least $1 - \epsilon$,
\begin{align*}
 \frac{|\mathcal{M}|}{N} \leq \frac{c_0 K P \delta^2 \log (8N/\epsilon)}{N\lambda_K^2}.
\end{align*}
\end{theorem}

The asymptotics of the mis-clustering rate depend on the number of covariates and the sparsity of the graph. This is demonstrated by Corollary \ref{corR}, which provides insight into how the number of covariates and graph sparsity affect the mis-clustering rate and the choice of tuning parameter.

\begin{corollary}\label{corR} (Mis-clustering bound with increasing $R$)
Assume $B_{i,i} = p, \forall i$ and $B_{i,j} = q, \forall i \neq j$; in addition, $M_{i,i} = m_1, \forall i$; $M_{i,j} = m_2, \forall i \neq j$; and $R>1$. For computational convenience, assume that each block has the same number of nodes $N/K$ and $R$ is a multiple of $K$, where $K$ is fixed. Let $R = \Theta\{(\log N)^{a+1}\}$, $d+\tau = \Theta\{(\log N)^{b+1}\}$, and $\alpha = \Theta\{N^{-1}(\log N)^{-1-c}\}$, where $a, b, c \geq 0$, then the mis-clustering bound from Theorem \ref{mcRate} becomes

\begin{align*}
\frac{|\mathcal{M}|}{N} \leq c_2\frac{(\log N)^{a-2c}+(\log N)^{(a-b)/2-c}+(\log N)^{-b}}{(\log N)^{2(a-c)}+(\log N)^{a-c}+\Theta(1)}.
\end{align*}

When $a \leq b$ the mis-clustering rate is minimized with $c=\frac{a+b}{2}$, which yields a rate of $O\{(\log N)^{-b}\}$. When $a>b$ the mis-clustering rate is minimized with $c=0$, which gives a rate of $O\{(\log N)^{-a}\}$.  
\end{corollary}

These results demonstrate that the tuning parameter $\alpha$ is determined by the balance between the number of covariates and the sparsity of the graph. A non-zero optimal $\alpha$ value signifies that including covariates does improve the mis-clustering bound, although it might not improve the asymptotics of the bound. Furthermore, the asymptotic mis-clustering rate is determined by the asymptotic behavior of the number of covariates or the mean number of edges, whichever is greater as determined by $a$ and $b$, respectively. For example, if we allow the number of covariates to grow with the number of nodes such that $a=0$ or $R = \Theta(\log N)$ and let the mean number of edges increase such that $b=0$ or $d+\tau = \Theta(\log N)$, then the covariates and the graph contribute equally to the asymptotic mis-clustering rate.

\begin{remark}\label{remarkh}
It is instructive to compare the value of $\alpha$ suggested by the results in Corollary \ref{corR} with the possible values of $\alpha$ based on the optimization procedure in \S \ref{oProc}. Computing $\alpha_{\rm min}$ and $\alpha_{\rm max}$ with $\tilde{\mathcal{L}}$, instead of $\tilde{L}$, for convenience, gives $\alpha_{\rm min} = \Theta\{(NR)^{-1}\}$ and $\alpha_{\rm max} = \Theta\{(NR)^{-1}\}$. Therefore, the optimization procedure will yield $\alpha = \Theta\{(NR)^{-1}\} = \Theta\{(N)^{-1}(\log N)^{-a-1}\}$. This agrees with the results of Corollary \ref{corR} when the mean node degree and the number of covariates grow at the same rate with respect to the number of nodes or $a = b$.

\end{remark}

\begin{corollary}\label{corr2}(Perfect clustering with increasing $R$)
Based on Theorem \ref{mcRate}, perfect clustering requires $\delta \{c_0KP\log(8N/\epsilon)\}^{1/2} < \lambda_K$. Under the simplifying assumptions given in Corollary \ref{corR}, perfect clustering is achieved when the number of covariates is $R \geq \Theta(N \log N)$.
\end{corollary}

\subsection{General lower bound}
The next theorem gives a lower bound for clustering a graph with node covariates. This bound uses Fano's inequality and is similar to that shown in \cite{chaudhuri2012spectral} for a graph without node attributes. We restrict ourselves to a node-contextualized stochastic blockmodel with $K = 2$ blocks, but allow an arbitrary number of covariates $R$.
\begin{theorem}\label{lowerBound}
(Covariate-assisted clustering lower bound) Consider the node-contextualized stochastic blockmodel with $K = 2$ blocks and $B$ such that $B_{1,1} \geq B_{2,2} \geq B_{1,2}$. Let the Kullback--Leibler divergence of the covariates be $\Gamma = \sum^R_{i=1} KL(\gamma_i, \gamma'_i)$, where $\gamma_i$ and $\gamma'_i$ are the distribution of the $i$th covariate under opposite block assignments, and $\Delta = B_{1,1} - B_{1,2}$. For a fixed $B_{1,1}$ and $N \geq 8$, in order to correctly recover the block assignments with probability at least $1 - \epsilon$, $\Delta$ must satisfy
\begin{align*}
\Delta \geq \frac{B_{1,1}(1-B_{1,1})}{\left[\frac{2}{N} \left\{\frac{\log 2}{2}(1-\epsilon) - \Gamma - \frac{\log 2}{N} \right\}\right]^{-1/2}+(1-B_{1,1})}.
\end{align*}
\end{theorem}

\begin{remark} (Lower bound interpretation)
If
\begin{align}
B_{1,1} - B_{1,2} &< \frac{B_{1,1}(1-B_{1,1})N^{-1/2}}{\left\{ \left(1-\epsilon - \frac{2}{N} \right) \log 2 \right\}^{-1/2}+(1-B_{1,1})} 
\text{, } \label{lBEq1} \\
\Gamma &< \left( \frac{1}{2} - \frac{\epsilon}{2} - \frac{1}{N} \right) \log 2, \label{lBEq2}
\end{align}
then only an algorithm that uses both the graph and node covariates can yield correct blocks with high probability. Condition \eqref{lBEq1} specifies when the graph is insufficient and condition \eqref{lBEq2} specifies when the covariates are insufficient to individually recover the block membership with high probability.
\end{remark}

\begin{remark} (Bound comparison)
The upper bound for covariate-assisted spectral clustering in Theorem \ref{mcRate} can be compared to the general lower bound. Simplifying the general lower bound gives the condition $\Delta \geq \Theta(N^{-1/2})$ for perfect clustering with probability $1 - \epsilon$. This is the same condition as for regularized spectral clustering. According to Theorem \ref{mcRate}, for this method to achieve prefect clustering with probability $1 - \epsilon$ requires $\delta \{c_0 K P \log(8N/\epsilon)\}^{1/2} < \lambda_K$. As highlighted in Corollary \ref{corr2} this condition cannot be satisfied for a fixed $R$, so it cannot be shown that covariate-assisted spectral clustering achieves perfect clustering for a fixed number of covariates. This is consistent with similar results for regularized spectral clustering. 
\end{remark} 

\section{SIMULATIONS} \label{simSection}
\subsection{Varying graph or covariate signal}
In these simulations, consider a node-contextualized stochastic blockmodel with $K = 3$ blocks and $R = 3$ node Bernoulli covariates. Define the block probabilities for the assortative graph, the non-assortative graph, and the covariates as
\begin{align}\label{simProb}
	B = \begin{bmatrix}
	p & q & q \\
	q & p & q \\
	q & q & p \\
	\end{bmatrix},&       & 
	B' = \begin{bmatrix}
	q & p & p \\
	p & q & p \\
	p & p & q \\
	\end{bmatrix},&       & 
	M = \begin{bmatrix}
	m_1 & m_2 & m_2\\
	m_2 & m_1 & m_2\\
	m_2 & m_2 & m_1\\
	\end{bmatrix},
\end{align}
where $p > q$ and $m_1 > m_2$. This implies that for the assortative graph the probability of an edge within a block is $p$, which is greater than $q$, the probability of an edge between two blocks. The opposite is true for the non-assortative graph. In the $k$th block, the probability of the $k$th covariate being one is $m_1$ and the probability of the other covariates being one is $m_2$.

These simulations compare five different methods. The first three are canonical correlation analysis clustering, covariate-assisted spectral clustering, and assortative covariate-assisted spectral clustering, which utilize the node edges as well as the node covariates to cluster the graph. The other two methods utilize either the node edges or the node covariates. For the node edges, regularized spectral clustering is used; for the node covariates, spectral clustering on the covariate matrix is used.
 
The first set of simulations investigates the effect of varying the block signal in the graph on the mis-clustering rate. This is done by varying the difference in the within and between block probabilities, $p-q$. The simulations are conducted for the assortative and non-assortative graphs, using $B$ and $B'$ in \ref{simProb}, shown in Fig. \ref{simFig}(a) and (b), respectively. In the assortative case, our assortative clustering method performs better than any of the other methods. Covariate-assisted spectral clustering performs slightly worse than the assortative variant, but still outperforms the other methods. In the non-assortative case, our clustering method has the best performance, while the assortative version always does worse than using only the covariates or the graph.
\begin{figure}[!h]
	\figurebox{35pc}{}{}[plotBiom3.eps]
	\caption{Average mis-clustering rate of five different clustering methods: covariate-assisted spectral clustering (solid), assortative covariate-assisted spectral clustering (dash), canonical correlation analysis clustering (dot), regularized spectral clustering (dot-dash), and spectral clustering on the covariate matrix (long dash). The fixed parameters are $N = 1500$, $p = 0.03$, $q = 0.015$, $m_1 = 0.8$, and $m_2 = 0.2$.}\label{simFig}
\end{figure}

The second set of simulations investigates the effect of varying the block signal of the covariates on the mis-clustering rate by changing the difference between the block specific covariate probabilities, $m_1 - m_2$. As shown in Fig. \ref{simFig}(c), assortative covariate-assisted spectral clustering tends to have a better mis-clustering rate than the other methods. Only when the difference in the covariate block probabilities is very small and $X$ effectively becomes a noise term does regularized spectral clustering outperform our assortative clustering method. For the non-assortative case shown in Fig. \ref{simFig}(d), assortative covariate-assisted spectral clustering performs poorly, while covariate-assisted spectral clustering is able to outperform all other methods for a sufficiently large difference in the covariate block probabilities. This is expected since the covariates in the assortative variant effectively increase the edge weights within a block, which will smooth out the block structure specified by $B'$.

\subsection{Model misspecification}
The final simulation considers the case where the block membership in the covariates is not necessarily the same as the block membership in the graph. The node Bernoulli covariates no longer satisfy \eqref{covModel} in Definition \ref{SBM}, but $\mathcal{X} = Y M$, where $Y \in \{0, 1\}^{N \times K}$ is a block membership matrix that differs from $Z$. As such, the underlying clusters in the graph do not align with the clusters in the covariates. This simulation varies the proportion of block assignments in $Y$ which agree with the block assignments in $Z$ to investigate the robustness of the methods to this form of model misspecification. The results in Fig. \ref{simFig}(e) show that assortative covariate-assisted spectral clustering is robust to covariate block membership model misspecification for the assortative graph. The mis-clustering rate shown is computed relative to the block membership of the graph. For this specific case, our assortative clustering method is able to achieve a lower mis-clustering rate than regularized spectral clustering as long as the proportion of agreement between the block membership of the graph and the covariates is greater than 0.7. Since a three block model is used, the lowest proportion of agreement possible is one third due to identifiability. For the non-assortative graph, Fig. \ref{simFig}(f), covariate-assisted spectral clustering requires a slightly higher level of agreement at 0.8.  

\section{CLUSTERING DIFFUSION MRI CONNECTOME GRAPHS}\label{dataSection}
Assortative covariate-assisted spectral clustering was applied to brain graphs recovered from diffusion magnetic resonance imaging \citep{craddock2013imaging}. Each node in a brain graph corresponds to a voxel in the brain. The edges between nodes are weighted by the number of estimated fibers that pass through both voxels. The center of a voxel is treated as the spatial location of the corresponding node. These spatial locations were centered and used as the first set of covariates in the analysis. The data set used in this analysis contains 42 brain graphs obtained from 21 different individuals. Only the largest connected components of the brain graphs were used, ranging in size from 707,000 to 935,000 nodes, with a mean density of 744 edges per node. In addition, the brain graphs contain brain atlas labels corresponding to 70 different neurological brain regions, which were treated as a second set of covariates.
\begin{figure}[!h]
	\figurebox{20pc}{}{}[brainPlotBiom_flat.eps]
	\caption{A section of a brain graph with nodes plotted at their spatial location and colored by cluster membership for three different clustering methods and a brain atlas.}\label{brainGraphFig}
\end{figure}

Whereas the simulations attempted to demonstrate the effectiveness of our clustering method in utilizing node covariates to help discover the underlying block structure of the graph, this analysis focuses on the ability of our clustering method to discover highly connected clusters with relatively homogeneous covariates. The node covariates contextualize the brain clusters and improve their interpretability. Like other clustering methods, covariate-assisted spectral clustering is mainly an exploratory tool which may or may not provide answers directly but can often provide insight about relationships within the data. In this example, it is used to examine the relationships between brain graph connectivity, spatial location, and brain atlas labels. 
The utility of covariate-assisted spectral clustering was explored by partitioning the brain graphs into 100 clusters. The brain graphs in this data set are assortative, so our assortative clustering method was used in this analysis. Since the brain graphs have heterogeneous node degrees, the rows of the eigenvector matrix were normalized when applying the spectral clustering algorithm to improve the clustering results \citep{qin2013regularized}. Figure \ref{brainGraphFig} shows a section of a sample brain graph with nodes plotted at their corresponding spatial locations and colored by cluster membership. For reference, the neurological brain atlas clusters with 70 different regions and an additional category for unlabeled nodes are also plotted. The brain graphs were clustered using three different approaches: regularized spectral clustering, assortative covariate-assisted spectral clustering with spatial location and with brain atlas membership. The tuning parameter $\alpha$ was set using the procedure in \S \ref{oProc}, and the values were $\alpha = 0.0004$ with spatial location covariates and $\alpha = 0.0708$ with brain atlas membership covariates.

As shown in Fig. \ref{brainGraphFig}, regularized spectral clustering yielded spatially diffuse clusters of densely connected nodes. By adding spatial location using covariate-assisted spectral clustering, we obtained densely connected and spatially coherent clusters. Regularized spectral clustering had two clusters of about 80,000 nodes and 4 clusters with fewer than 1,000 nodes, while the largest cluster from our clustering method had fewer than 50,000 nodes and no clusters had fewer than 1,000 nodes. Both greater spatial coherence and increased uniformity in cluster size demonstrated by covariate-assisted spectral clustering are important qualities for interpreting the partition. In addition, the clusters have a greater similarity with the brain atlas labels, though this similarity is still not very substantial. This suggests that brain graph connectivity is governed by more than just the neurological regions in the brain atlas. 
\begin{table}[!h]
\def~{\hphantom{0}}
\tbl{The adjusted Rand index between different partitions}{
\begin{tabular}{lccccc}
& ACASC-X & Brain Atlas & ACASC-BA & SC-X \\
RSC &  0.095 & 0.082 & 0.085 & 0.092 \\
ACASC-X & - & 0.169 & 0.189 & 0.278\\
Brain Atlas & - & - & 0.838 & 0.226 \\
ACASC-BA & - & - & - & 0.227 \\
\end{tabular}} \label{AriTable}
\begin{tabnote}
    RSC, regularized spectral clustering; ACASC-X, assortative covariate-assisted spectral clustering with spatial location;  ACASC-BA assortative covariate-assisted spectral clustering with brain atlas membership
\end{tabnote}
\end{table}

The relation between the brain atlas and the brain graph was studied further by treating brain atlas membership as the node covariates. This allowed the discovery of highly-connected regions with relatively homogeneous graph atlas labels. As shown in Fig. \ref{brainGraphFig}, relative to the brain atlas, some of the clusters are broken up, a few are joined together, and others overlap with multiple brain atlas regions, but the high similarity is clearly visible. Importantly, this approach gives us clusters that are highly aligned with known neurological regions while allowing for individual variability of the partitions based on brain graph connectivity. The adjusted Rand index was used to quantify the similarity of the partitions of a brain graph specified by the different clustering methods and the brain atlas in Table \ref{AriTable}. The alignment with the partitions based only on spatial location and either covariate-assisted spectral clustering with spatial location or the brain atlas is greater than between the two methods. This indicates that both the clusters from our method and the brain atlas are spatially coherent yet not highly overlapping.
\begin{figure}[!h]
	\figurebox{10pc}{}{}[brainHeatmap.eps]
	\caption{Heat maps comparing the partitions of 42 brain graphs using the adjusted Rand index with adjacent rows corresponding to two scans of the same individual.}\label{heatMapFig}
\end{figure}
Brain graph connectivity appears to be giving the clusters that use spatial location a different configuration from the brain atlas, as seen in Fig. \ref{brainGraphFig}. As expected, covariate-assisted spectral clustering with brain atlas membership has the highest adjusted Rand index partition similarity with the brain atlas but low similarity with the regularized spectral clustering partitions. If a more balanced partition alignment is desired, the tuning parameter can be adjusted accordingly. 

The relationship between all 42 brain graphs was analyzed by using the adjusted Rand index to compare partitions between them, as shown in Fig. \ref{heatMapFig}. To conduct the comparison, the nodes of each brain graph were matched by spatial location, and any non-matching nodes were ignored. Both regularized spectral clustering and covariate-assisted spectral clustering with spatial location distinguish clearly between individuals based on their brain graph partitions, but the latter gave partitions which are more homogeneous both within and between individuals. This increased partition consistency is favorable since a high degree of variation in the clusters between individuals would make them more difficult to interpret. 

\section{DISCUSSION}	
	
	Although the node-contextualized stochastic blockmodel is useful for studying graph clustering methods, data can deviate from the model's assumptions. More generally, covariate-assisted spectral clustering can be used to find highly connected communities with relatively homogeneous covariates, where the balance between these two objectives is controlled by the tuning parameter and can be set empirically or decided by the analyst. Relatively homogeneous covariates contextualize the clusters, making them easier to interpret and allow the analyst to focus on partitions that align with important covariates. Beyond its scientific interest, the brain graph analysis demonstrates the computational efficiency of our clustering method since the analysis could not have been feasibly conducted with existing methods. Nevertheless, determining an optimal tuning parameter value still presents a computational burden. Using a low rank update algorithm for eigenvector decomposition can further reduce this cost. 
	
	This work is meant as a step towards statistical understanding of graphs with node covariates. Further work is needed to better understand the use of covariate-assisted spectral clustering for network contextualization. Methods for determining the relative contribution of the graph and the covariates to a graph partition and tests to signify which covariates are informative would be useful tools. Ultimately, a thorough examination of the relationship between graph structure and node covariates is essential for a deep understanding of the underlying social or biological system.  

\section*{ACKNOWLEDGMENT}
This research was supported by NIH grant 5T32HL083806-
08, NSF grants DMS-1309998, DMS-1612456, ARO grant W911NF1510423, the Defense Advanced Research Projects Agency (DARPA) SIMPLEX program through SPAWAR contract N66001-15-C-4041 and DARPA GRAPHS N66001-14-1-4028. The authors would like to thank Yilin Zhang, Tai Qin, Jun Tao, Soumendu Sundar Mukherjee, and Zoe Russek for helpful comments.

\section*{SUPPLEMENTARY MATERIAL}
\label{SM}
\subsection{Discontinuous Transitions in the Leading Eigenspace of $\tilde{L}$}\label{tuning}
Discontinuous changes in the leading eigenspace of $\tilde{L}(\alpha)$ are a major concern when determining an optimal $\alpha$ value since they have a large effect on the clustering results. They can be studied algebraically by expressing $\tilde{L}(\alpha)$ in terms of the eigenvectors of $L_{\tau}L_{\tau}$ and $XX^T$. This approach is motivated by \cite{brand2006fast}.

Let $L_{\tau}L_{\tau} = V \Lambda V^T$ and $P$ be the orthogonal basis of the column space of $(I-VV^T)XX^T$, the component of $XX^T$ orthogonal to $V$. Let $XX^T = \tilde{V} \tilde{\Lambda} \tilde{V}^T$ and $\tilde{X}_i = \tilde{\lambda}_i^{1/2} \tilde{V}_i$, so $XX^T = \tilde{X}\tilde{X}^T$. Then, $\tilde{L}$ can be written as follows. 
\begin{align*}
\tilde{L} &= L_{\tau}L_{\tau} + \alpha XX^T \\
&= L_{\tau}L_{\tau} + \alpha \tilde{X}\tilde{X}^T \\
&= 
\begin{bmatrix}
V & \tilde{X}
\end{bmatrix}
\begin{bmatrix}
\Lambda & 0 \\
0 & \alpha I \\
\end{bmatrix}
\begin{bmatrix}
V & \tilde{X}
\end{bmatrix}^T \\
&= 
\begin{bmatrix}
V & P
\end{bmatrix}
\begin{bmatrix}
I & V^T\tilde{X} \\
0 & P^T(I-VV^T)\tilde{X} \\
\end{bmatrix}
\begin{bmatrix}
\Lambda & 0 \\
0 & \alpha I \\
\end{bmatrix}
\begin{bmatrix}
I & 0 \\
\tilde{X}^TV & \tilde{X}^T(I-VV^T)P \\
\end{bmatrix}
\begin{bmatrix}
V & P
\end{bmatrix}^T \\
&=
\begin{bmatrix}
V & P
\end{bmatrix}
\begin{bmatrix}
\Lambda + \alpha V^T\tilde{X}\tilde{X}^TV & \alpha V^T\tilde{X}\tilde{X}^T(I-VV^T)P \\
\alpha P^T(I-VV^T)\tilde{X}\tilde{X}^TV & \alpha P^T(I-VV^T)\tilde{X}\tilde{X}^T(I-VV^T)P \\
\end{bmatrix}
\begin{bmatrix}
V & P
\end{bmatrix}^T \\
&= 
\begin{bmatrix}
V & P
\end{bmatrix}
S
\begin{bmatrix}
V & P
\end{bmatrix}^T \\
&= \Big(
\begin{bmatrix}
V & P
\end{bmatrix}
V'\Big)\Lambda'\Big(V'^T
\begin{bmatrix}
V & P
\end{bmatrix}^T\Big).
\end{align*}

Note that 
\begin{align*}
(\Lambda + \alpha V^T\tilde{X}\tilde{X}^TV)_{ij} = \lambda_i \delta_{ij} + \alpha \sum_k (V_i^T\tilde{X}_k) (\tilde{X}_k^TV_j) 
\end{align*}
and
\begin{align*}
P^T(I-VV^T)\tilde{X}\tilde{X}^TV = \{P^T(I-VV^T)\tilde{X}\}[(\tilde{X}_i^TV_j)_{ij}].
\end{align*}
Hence, for any $j$ such that $\tilde{X}_i^TV_j = 0, \forall i$, the $j$th row and column of $S$ will be zero except for the diagonal element. This means that $U_j$ will not be rotated by $V'$ and will be an eigenvector of $\tilde{L}$ for all values of $\alpha$. The eigenvalue $\lambda_j$ will not change either, but its position relative to the other eigenvalues will change with $\alpha$. The change in the relative position of $\lambda_j$ will result in a discontinuous transition in the leading eigenspace of $\tilde{L}$ if $j \geq K$.

For any $i$ such that $\tilde{X}_i^TV_j = 0, \forall j$, $\tilde{V}_i$ is a column in $P$ by construction. Row $i$ in the lower left block of $S$ is
\begin{align*}
\tilde{V}_i^T(I-VV^T)\tilde{X}[(\tilde{X}_i^TV_j)_{ij}] &= [0,...,\tilde{\lambda}_i^{1/2}, 0, ...][(\tilde{X}_i^TV_j)_{ij}] \\
& \qquad -  [0,..., 1, 0, ...] \text{diag}\Big(\tilde{\lambda}_1^{1/2}, ..., \tilde{\lambda}_R^{1/2}\Big)[(\tilde{X}_i^TV_j)_{ij}]\\
&= [0, ..., 0] ,
\end{align*}
and, since $S$ is symmetric, this is also column $i$ in the upper right block of $S$. The lower right block of $S$ has row $i$, and by symmetry column $i$, given by 
\begin{align*}
\tilde{V}_i^T(I-VV^T)\tilde{X}\tilde{X}^T(I-VV^T)P &= \tilde{v}_i^T(\tilde{X}\tilde{X}^T - \tilde{X}\tilde{X}^TVV^T)P \\
&= \tilde{\lambda}_i \tilde{V}_i^T P \\
&= [0, ..., \tilde{\lambda}_i, 0, ...] .
\end{align*}
Thus, for any $i$ such that $\tilde{X}_i^TV_j = 0, \forall j$ the $i$th row and column of $S$ will be zero except for the diagonal element. This means that $\tilde{V}_i$ and $\tilde{\lambda}_i$ will be an eigenvector and eigenvalue of $\tilde{L}$ for all values of $\alpha$, but will occupy different relative positions in the eigendecomposition based on the value of $\alpha$. The change in the relative position of $\tilde{\lambda}_i$ will result in a discontinuous transition in the leading eigenspace of $\tilde{L}$ if $i \geq K$.

Knowing the interval on which such discontinuous transitions are possible can reduce the computational burden of choosing an optimal $\alpha$. The values of $\alpha$ for which transitions occur can be identified as points at which the eigengap equals zero, $\lambda_{K}(\tilde{L}) - \lambda_{K+1}(\tilde{L}) = 0$. First, consider the lowest possible value of $\alpha$ for which such a transition can occur, $\alpha = \text{argmin}_{\alpha} \{\alpha: \lambda_{K}(\tilde{L}) - \lambda_{K+1}(\tilde{L}) = 0\}$. Note that $\lambda_K(\tilde{L}) \geq \lambda_K(L_{\tau}L_{\tau})$, where the equality holds when $V_K$ is orthogonal to $X$ and $\alpha$ is sufficiently small, and $\lambda_{K+1}(\tilde{L}) \leq \lambda_{K+1}(L_{\tau}L_{\tau}) + \alpha \lambda_1(XX^T)$, where the equality holds when $V_{K+1}$ is identical to $\tilde{V}_1$. Hence, the earliest possible transition occurs when
\begin{align*}
\lambda_{K}(L_{\tau}L_{\tau}) &- \{\lambda_{K+1}(L_{\tau}L_{\tau}) + \alpha_{min} \lambda_1(XX^T)\} = 0, \\
\alpha_{min} &= \frac{\lambda_{K}(L_{\tau}L_{\tau}) - \lambda_{K+1}(L_{\tau}L_{\tau})}{\lambda_1(XX^T)}.
\end{align*} 
For the highest value of $\alpha$ for which such a transition is possible, consider $\alpha^{-1}\tilde{L}$. Following the above argument for $\alpha^{-1}$ with $XX^T$ and $L_{\tau}L_{\tau}$ interchanged, a symmetric result is obtained with the additional dependence on the number of covariates, $R$. This result yields,  
\begin{align*}
\alpha_{max} &= \frac{\lambda_{1}(L_{\tau}L_{\tau})}{\lambda_{R}(XX^T) \mathbf{1}_{(R \leq K)} + \{\lambda_{K}(XX^T) - \lambda_{K+1}(XX^T)\} \mathbf{1}_{(R > K)}} .
\end{align*}
Therefore, discontinuous transitions in the leading eigenspace of $\tilde{L}(\alpha)$ can only occur in the interval $[\alpha_{\rm min}, \alpha_{\rm max}]$.

\subsection{Empirical Results for Choosing $\alpha$}
Figure \ref{suppSim} presents some empirical details to demonstrate how the within cluster sum of squares and the mis-clustering rate vary with the tuning parameter $\alpha$. The simulations shown in the figure use the same model structure described in \S 4 of the paper. The results show the minimum of the within cluster sum of squares falls within the prescribed range of $\alpha$, $[\alpha_{\rm min}, \alpha_{\rm max}]$. Furthermore, the minimum of the within cluster sum of squares tends to align with the minimum of the mis-clustering rate. Similar results were observed for other parameter settings. 
\begin{figure}[!h]
	\figurebox{16pc}{}{}[suppSim.eps]
	\caption{The results of assortative covariate-assisted spectral clustering for a range of $\alpha$ values. The solid line in bottom graphs indicates the $\alpha$ value chosen by the optimization procedure and the dased lines indicate the interval $[\alpha_{\rm min}, \alpha_{\rm max}]$. The fixed parameters are $N = 1500$, $p = 0.03$, $m_1 = 0.8$, and $m_2 = 0.2$.}\label{suppSim}
\end{figure}

\subsection{Proof of Lemma 1}\label{pfBlockLemma}
This proof follows the approach used in \cite{rohe2011spectral} to establish the equivalence between block membership and a subset of the population eigenvectors. Note that $\tilde{\mathcal{L}} = (\mathcal{D} + \tau I)^{-1/2} Z B Z^T (\mathcal{D} + \tau I)^{-1} Z B Z^T (\mathcal{D} + \tau I)^{-1/2} + \alpha E(XX^{T})$. Define $c_l = \sum_i Var(X_{il}|Z_i=l)$, a diagonal matrix $\tilde{C}$ such that $\tilde{C}_{ll} = c_l$, and a diagonal matrix $C$ such that $CZ = Z\tilde{C}$. 

If we let $\mathcal{D}_B = \text{diag}(BZ^T\mathbf{1}_n + \tau)$, then 
$\tilde{\mathcal{L}} = Z\{\mathcal{D}_B^{-1/2} B Z^T (\mathcal{D} + \tau I)^{-1} Z B \mathcal{D}_B^{-1/2} + \alpha M M^T\} Z^T + \alpha C$. Recall that $B$ is symmetric and full rank by assumption.
Let $\tilde{B} = \mathcal{D}_B^{-1/2} B Z^T (\mathcal{D} + \tau I)^{-1} Z B \mathcal{D}_B^{-1/2} + \alpha M M^T$, which is positive definite $\forall \alpha \ge 0$. Assume $\alpha$ is chosen such that $\tilde{B}$ is full rank, which is true $\forall \alpha$ with the possible exception of a set of values of measure zero. Let $\tilde{P}=Z^T Z$ and note that $\det(\tilde{B}\tilde{P})=\det(\tilde{B})\det(\tilde{P}) > 0$. Hence, $\tilde{B}\tilde{P} + \alpha \tilde{C}$ is symmetric and has real eigenvalues. By spectral decomposition, let
\begin{align*}
\tilde{B}\tilde{P}+ \alpha \tilde{C} = \mu \Lambda \mu^T .
\end{align*}
Then,
\begin{align*}
\tilde{\mathcal{L}}Z\mu &= (Z\tilde{B}Z^T + \alpha C)Z\mu \\
&= (Z\tilde{B}Z^TZ + \alpha CZ)\mu \\
&= (Z(\tilde{B}\tilde{P}) + \alpha Z\tilde{C})\mu \\
&= Z \mu \Lambda .
\end{align*}

Therefore, $Z\mu$ is the matrix of $K$ eigenvectors of $\tilde{\mathcal{L}}$, but not necessarily the top $K$. Also, $\det(\mu)>0$ so $\mu^{-1}$ exists and $Z_i\mu = Z_j\mu \iff Z_i = Z_j$.
This establishes the equivalence between block membership and a subset of the population eigenvectors. A condition will now be derived for which this equivalence holds for the top $K$ population eigenvectors.
Let $x$ be a normalized eigenvector orthogonal to the span of $Z\mu$. Because $\mu$ has orthogonal columns, it is full rank. As such, $x^TZ = 0$.

Define $\bar{c} = \sum^K c_l/K$, $\bar{C} = \bar{c}I$, and $\varkappa = \max_l |c_l-\bar{c}|$, then 

\begin{align*}
x^{T} \tilde{\mathcal{L}} x &= x^{T} (Z\tilde{B}Z^{T} + \alpha C) x \\
&= \alpha x^{T} C x \\
&= \alpha x^{T}(\bar{C} + (C - \bar{C}))x \\
&= \alpha x^{T}\bar{c}I x + \alpha x^{T}(C-\bar{C})x \\
&= \alpha \bar{c} + \alpha x^{T}(C - \bar{C})x \\
&\leq \alpha \bar{c} + \alpha ||C - \bar{C}|| \\
&= \alpha(\bar{c} + \varkappa) .
\end{align*}

The $k$th eigenvalue of $\tilde{B}\tilde{P}+\alpha \tilde{C}$ is given by

\begin{align*}
\lambda_K(\tilde{B}\tilde{P}+\alpha \tilde{C}) &= \min_{||x||=1} x^{T} (\tilde{B}\tilde{P}+\alpha \tilde{C})x \\
&= \min x^{T} [(\tilde{B}\tilde{P}+ \alpha \bar{c}I)+(\alpha \tilde{C}-\alpha \bar{c}I)]x \\
&\geq \min x^{T} (\tilde{B}\tilde{P}+\alpha \bar{c}I)x + \alpha \min x^{T} (\tilde{C}-\bar{c}I)x \\
&= \min x^{T}\tilde{B}\tilde{P}x + \alpha \bar{c} - \alpha \max x^{T} (\bar{c}I - \tilde{C})x \\
&\geq \lambda_K(\tilde{B}\tilde{P}) + \alpha \bar{c} - \alpha \max_u |c_u-\bar{c}| \\
&= \lambda_K(\tilde{B}\tilde{P}) + \alpha (\bar{c} - \varkappa).
\end{align*}

Hence, a positive eigengap exists between the eigenvectors in $Z\mu$ and $x$ if
\begin{align*}
 0 &< \lambda_K(\tilde{B}\tilde{P}+\alpha \tilde{C}) - \max_x x^{T} (Z\tilde{B}Z^{T} + \alpha C) x \\
&<  \lambda_K(\tilde{B}\tilde{P}) + \alpha (\bar{c} - \varkappa) - \alpha(\bar{c} + \varkappa) \\
&= \lambda_K(\tilde{B}\tilde{P}) - 2 \alpha \varkappa.
\end{align*}

Assume $(i)$ $\lambda_K(\tilde{B}\tilde{P}) > 2 \alpha \varkappa$, then the top $K$ eigenvectors of $\tilde{\mathcal{L}}$ are given by $Z\mu$, where $Z_i\mu = Z_j\mu \iff Z_i = Z_j$. Hence, there is an equivalence between block membership and the top $K$ population eigenvectors.

\subsection{Proof of Theorem 1}\label{pfConcentration}
\subsubsection{Triangle inequality bound}
The spectral norm of the difference between the sample and population covariate-assisted Laplacians is bounded by first applying the triangle inequality and bounding the resulting terms individually.
\begin{align}
 \| \tilde{L} - \tilde{\mathcal{L}} \| 
 &\leq \| \alpha XX^T - E(\alpha XX^T) \| \label{term1} \\ 
                         &+ \| \mathcal{D}_{\tau}^{-1/2} A \mathcal{D}_{\tau}^{-1} A \mathcal{D}_{\tau}^{-1/2} - \mathcal{D}_{\tau}^{-1/2} \mathcal{A} \mathcal{D}_{\tau}^{-1} \mathcal{A} \mathcal{D}_{\tau}^{-1/2} \| \label{term3} \\
                         &+ \| D_{\tau}^{-1/2} A D_{\tau}^{-1} A D_{\tau}^{-1/2} -\mathcal{D}_{\tau}^{-1/2} A \mathcal{D}_{\tau}^{-1} A \mathcal{D}_{\tau}^{-1/2} \| \label{term4}.
\end{align}

\subsubsection{Bound for Equation (\ref{term1})}
 For equation (\ref{term1}), use the matrix Bernstein inequality \citep{tropp2012user}. Note that $\alpha XX^T =\sum_k \alpha X_k X_k^T$, where $X_k$ is the $k$th column of $X$. Now bound the spectral norm of $\alpha X_k X_k^T-E(\alpha X_k X_k^T)$.
\begin{align*}
 \|  \alpha X_k X_k^T-E(\alpha X_k X_k^T) \| &= \alpha \| X_k X_k^T - \mathcal{X}_k \mathcal{X}_k^T - \text{diag}(\mathcal{X}_k^{(2)} - \mathcal{X}_k^2) \| \\
 & \leq \alpha ( \| X_k X_k^T \| + \| \mathcal{X}_k \mathcal{X}_k^T \| + \max |\mathcal{X}_k^{(2)} - \mathcal{X}_k^2 |) \\
&\leq \alpha (N J^2 + N J^2 + J^2) \\
&\leq 3 \alpha NJ^2 \\
& \equiv S.
\end{align*}
Next, find a bound on the spectral norm of the variance of $\alpha XX^T$. Let $\mathcal{X}_k^{(i)}$ be the $i$th moment of $X_k$. Note that vector products are element-wise where dictated by vector dimensions. 
\begin{align*}
E(X_k X_k^T) =& \mathcal{X}_k \mathcal{X}_k^T - \text{diag}(\mathcal{X}_k^2 - \mathcal{X}_k^{(2)}).\\
E(X_k X_k^T) E(X_k X_k^T) =& \{\mathcal{X}_k \mathcal{X}_k^T - \text{diag}(\mathcal{X}_k^2 - \mathcal{X}_k^{(2)})\}
\{\mathcal{X}_k \mathcal{X}_k^T - \text{diag}(\mathcal{X}_k^2 - \mathcal{X}_k^{(2)})\}\\
=& \mathcal{X}_k \mathcal{X}_k^T \mathcal{X}_k \mathcal{X}_k^T - \mathcal{X}_k \mathcal{X}_k^T \text{diag}(\mathcal{X}_k^2 - \mathcal{X}_k^{(2)}) \\
&- \text{diag}(\mathcal{X}_k^2 - \mathcal{X}_k^{(2)}) \mathcal{X}_k \mathcal{X}_k^T + \text{diag}\{(\mathcal{X}_{k}^2 - \mathcal{X}_{k}^{(2)})^2\} \\
=& \Big(\sum_i \mathcal{X}_{ik}^2\Big) \mathcal{X}_k \mathcal{X}_k^T - \mathcal{X}_k \{\mathcal{X}_k (\mathcal{X}_k^2 - \mathcal{X}_k^{(2)})\}^T \\
&- \{\mathcal{X}_k (\mathcal{X}_k^2 - \mathcal{X}_k^{(2)})\} \mathcal{X}_k^T + \text{diag}\{(\mathcal{X}_{k}^2 - \mathcal{X}_{k}^{(2)})^2\}. 
\end{align*}
\begin{align*}
E(X_k X_k^T X_k X_k^T) =& E\Big\{\Big(\sum_i X_{ik}^2\Big)X_k X_k^T\Big\}\\
=& \begin{cases}
\mathcal{X}_{ik} \mathcal{X}_{jk} \sum_{l \neq i,j} \mathcal{X}_{lk}^{(2)} + \mathcal{X}_{ik} \mathcal{X}_{jk}^{(3)} + \mathcal{X}_{jk} \mathcal{X}_{ik}^{(3)} & i \neq j \\
\mathcal{X}_{ik}^{(2)} \sum_{l \neq i} \mathcal{X}_{lk}^{(2)} + \mathcal{X}_{ik}^{(4)} & i = j
\end{cases}\\
=& \Big(\sum \mathcal{X}_{ik}^{(2)}\Big)\mathcal{X}_k \mathcal{X}_k^T - \mathcal{X}_k (\mathcal{X}_k \mathcal{X}_k^{(2)})^T - (\mathcal{X}_k \mathcal{X}_k^{(2)}) \mathcal{X}_k^T \\
&+ \mathcal{X}_k \mathcal{X}_k^{(3)^T} + \mathcal{X}_k^{(3)} \mathcal{X}_k^T \\
&+ \text{diag}\Big\{(\mathcal{X}_k^{(2)}-\mathcal{X}_k^2)\Big(\sum_i \mathcal{X}_{ik}^{(2)}\Big) - \mathcal{X}_k^{(2)^2} + 2 \mathcal{X}_k^2 \mathcal{X}_k^{(2)} - 2 \mathcal{X}_k \mathcal{X}_k^{(3)} + \mathcal{X}_k^{(4)}\Big\}.
\end{align*}
\begin{align*}
\text{Var}(X_k X_k^T) =& \mathcal{X}_k \mathcal{X}_k^T \sum_i (\mathcal{X}_{ik}^{(2)} - \mathcal{X}_{ik}^2)+ \mathcal{X}_k \{\mathcal{X}_k (\mathcal{X}_k^2 - 2\mathcal{X}_k^{(2)}) + \mathcal{X}_k^{(3)}\}^T +  \{\mathcal{X}_k (\mathcal{X}_k^2 - 2\mathcal{X}_k^{(2)}) + \mathcal{X}_k^{(3)}\} \mathcal{X}_k^T \\
&+ \text{diag} \left\{ (\mathcal{X}_k^{(2)}-\mathcal{X}_k^2)\Big(\sum_i \mathcal{X}_{ik}^{(2)}\Big) - \mathcal{X}_k^{(2)^2} + 2 \mathcal{X}_k^2 \mathcal{X}_k^{(2)} - 2 \mathcal{X}_k \mathcal{X}_k^{(3)} + \mathcal{X}_k^{(4)} - (\mathcal{X}_{k}^{(2)} - \mathcal{X}_{k}^2)^2 \right\}.
\end{align*}
\begin{align*}
\Big| \Big| \sum_k \text{Var}( X_k X_k^T) \Big| \Big| &\leq \sum_k \sum_i \Big| \mathcal{X}_{ik}^2 \sum_l ( \mathcal{X}_{lk}^{(2)} - \mathcal{X}_{lk}^2)| + 2 | \mathcal{X}_{ik}^4 - 2 \mathcal{X}_{ik}^2 \mathcal{X}_{ik}^{(2)} + \mathcal{X}_{ik} \mathcal{X}_{ik}^{(3)}\Big| \\
&+ \max_i \Big|(\mathcal{X}_{ik}^{(2)}-\mathcal{X}_{ik}^2)\Big(\sum_l \mathcal{X}_{lk}^{(2)}\Big) - \mathcal{X}_{ik}^{(2)^2} + 2 \mathcal{X}_{ik}^2 \mathcal{X}_{ik}^{(2)} - 2 \mathcal{X}_{ik} \mathcal{X}_{ik}^{(3)} + \mathcal{X}_{ik}^{(4)} - (\mathcal{X}_{ik}^{(2)} - \mathcal{X}_{ik}^2)^2 \Big| \\
&\leq \sum_k \sum_i \mathcal{X}_{ik}^2 \sum_l ( \mathcal{X}_{lk}^{(2)} - \mathcal{X}_{lk}^2) + 2(\mathcal{X}_{ik}^2 \mathcal{X}_{ik}^{(2)} - \mathcal{X}_{ik}^4) + 2| \mathcal{X}_{ik}^2 \mathcal{X}_{ik}^{(2)} - \mathcal{X}_{ik} \mathcal{X}_{ik}^{(3)}| \\
&+ \max_i \Big\{(\mathcal{X}_{ik}^{(2)}-\mathcal{X}_{ik}^2)\Big(\sum_l \mathcal{X}_{lk}^{(2)}\Big) + |2 \mathcal{X}_{ik} \mathcal{X}_{ik}^{(3)} - \mathcal{X}_{ik}^{(4)} - \mathcal{X}_{ik}^4| + 2(\mathcal{X}_{ik}^{(2)} - \mathcal{X}_{ik}^2)^2 \Big\}\\
&\leq \sum_k \sum_i 3 \mathcal{X}_{ik}^2 \sum_l ( \mathcal{X}_{lk}^{(2)} - \mathcal{X}_{lk}^2) + 2| \mathcal{X}_{ik}^2 \mathcal{X}_{ik}^{(2)} - \mathcal{X}_{ik} \mathcal{X}_{ik}^{(3)}| \\
&+ \max_i \Big\{3(\mathcal{X}_{ik}^{(2)}-\mathcal{X}_{ik}^2)\Big(\sum_l \mathcal{X}_{lk}^{(2)}\Big) + |2 \mathcal{X}_{ik} \mathcal{X}_{ik}^{(3)} - \mathcal{X}_{ik}^{(4)} - \mathcal{X}_{ik}^4|\Big\}\\
&\leq 8 \sum_k \left\{ \sum_i \mathcal{X}_{ik}^{(2)} \sum_l ( \mathcal{X}_{lk}^{(2)} - \mathcal{X}_{lk}^2) + \mathcal{X}_{ik}^{(4)} \right\} .
\end{align*}
Thus,
\begin{align*}
\Big| \Big| \sum_k \text{Var}(\alpha X_k X_k^T) \Big| \Big| \leq 8 \alpha^2 \sum_k \left\{ \sum_i \mathcal{X}_{ik}^{(2)} \sum_l ( \mathcal{X}_{lk}^{(2)} - \mathcal{X}_{lk}^2) + \mathcal{X}_{ik}^{(4)} \right\} \equiv \varpi .
\end{align*}

Let $b = \{3 \varpi \log(8N/\epsilon)\}^{1/2}$ and assume $(iii)$ $\varpi/S^2 > 3 \log(8N/\epsilon)$, then $b <  \varpi/S$. Note that assumption $(iii)$ requires that $R \geq \Theta(\log N)$. Applying the matrix Bernstein inequality gives,
\begin{align*}
  P( || \alpha XX^T - E(\alpha XX^T) || > b) &\leq 2N \exp \left( -\frac{b^2}{2\sigma^2+2 S b/3}\right)\\
&\leq 2N \exp \left\{ -\frac{3 S \varpi \log(8N/\epsilon)}{2 \varpi+2 S b/3}\right\}\\
&\leq 2N \exp \left\{ -\frac{3\varpi \log(8N/\epsilon)}{3\varpi}\right\}\\
&= \epsilon/4.
\end{align*}
Hence, with with probability $1-\epsilon/4$,
\begin{align*}
\| \alpha XX^T - E(\alpha XX^T) \| \leq b .
\end{align*}

\subsubsection{Bound for Equation (\ref{term3})}
Equation (\ref{term3}) can be decomposed into three terms using properties of the spectral norm. 
\begin{align*}
&\| \mathcal{D}_{\tau}^{-1/2} A \mathcal{D}_{\tau}^{-1} A \mathcal{D}_{\tau}^{-1/2} - \mathcal{D}_{\tau}^{-1/2} \mathcal{A} \mathcal{D}_{\tau}^{-1} \mathcal{A} \mathcal{D}_{\tau}^{-1/2} \|\\
& \leq
\| \mathcal{D}_{\tau}^{-1/2} A \mathcal{D}_{\tau}^{-1} A \mathcal{D}_{\tau}^{-1/2} - E(\mathcal{D}_{\tau}^{-1/2} A \mathcal{D}_{\tau}^{-1/2}) E(\mathcal{D}_{\tau}^{-1/2} A \mathcal{D}_{\tau}^{-1/2}) \| \\ 
& \leq 
\| \mathcal{D}_{\tau}^{-1/2} A \mathcal{D}_{\tau}^{-1/2} - E(\mathcal{D}_{\tau}^{-1/2} A \mathcal{D}_{\tau}^{-1/2}) \|
\| \mathcal{D}_{\tau}^{-1/2} A \mathcal{D}_{\tau}^{-1/2} + E(\mathcal{D}_{\tau}^{-1/2} A \mathcal{D}_{\tau}^{-1/2}) \|.
\end{align*}
The first term above can be bounded following the proof in the Supplement of \cite{qin2013regularized}. 
Under the assumption that $(ii)$ $d + \tau > 3 \log(8N/\epsilon)$, where $d=\min \mathcal{D}_{ii}$, let $a = [\{3\log(8N/\epsilon)\}/(d + \tau)]^{1/2}$, so $a < 1$. Then, with probability at least $1 - \epsilon/4$,
\begin{align*}
 \| \mathcal{D}_{\tau}^{-1/2} A \mathcal{D}_{\tau}^{-1/2} - E(\mathcal{D}_{\tau}^{-1/2} A \mathcal{D}_{\tau}^{-1/2}) \| \leq a .
 \end{align*}

Using the fact that $\|\mathcal{L}_{\tau} \| \leq 1$, $\|L_{\tau} \| \leq 1$, and $\| \mathcal{D}_{\tau}^{-1/2} D_{\tau}^{-1/2} \| \leq a + 1$, with probability $1-\epsilon/4$, as shown in the Supplement of \cite{qin2013regularized}, the second term can be bounded with probability $1-\epsilon/4$ as follows.
\begin{align*}
&\| \mathcal{D}_{\tau}^{-1/2} A \mathcal{D}_{\tau}^{-1/2} + E(\mathcal{D}_{\tau}^{-1/2} A \mathcal{D}_{\tau}^{-1/2}) \| \\
&\leq \| \mathcal{D}_{\tau}^{-1/2} D_{\tau}^{-1/2} L_{\tau} D_{\tau}^{-1/2} \mathcal{D}_{\tau}^{-1/2} \| + \| \mathcal{L}_{\tau} \| \\
&\leq \| \mathcal{D}_{\tau}^{-1/2} D_{\tau}^{-1/2} \| \| L_{\tau} \| \| D_{\tau}^{-1/2} \mathcal{D}_{\tau}^{-1/2} \| + 1 \\
&\leq (a+1)^2 + 1.
\end{align*}
Hence, with with probability $1-\epsilon/4$,
\begin{align*}
\| \mathcal{D}_{\tau}^{-1/2} A \mathcal{D}_{\tau}^{-1} A \mathcal{D}_{\tau}^{-1/2} - E(\mathcal{D}_{\tau}^{-1/2} A \mathcal{D}_{\tau}^{-1} A \mathcal{D}_{\tau}^{-1/2}) \|
 \leq a(a + 1)^2 + a .
\end{align*}

\subsubsection{Bound for Equation (\ref{term4})}
Note that $\| \mathcal{D}_{\tau}^{-1/2} D_{\tau}^{-1/2} - I \| \leq a$, with probability $1-\epsilon/4$, as shown in the Supplement of \cite{qin2013regularized}, and $\| D_{\tau}^{-1/2} \mathcal{D}_{\tau}^{-1} D_{\tau}^{-1/2} - I \| \leq a$, which can be derived by the same approach. Using these results, equation (\ref{term4}) can be bounded with probability $1-\epsilon/2$ as follows.
\begin{align*}
&\| D_{\tau}^{-1/2} A D_{\tau}^{-1} A D_{\tau}^{-1/2} -\mathcal{D}_{\tau}^{-1/2} A \mathcal{D}_{\tau}^{-1} A \mathcal{D}_{\tau}^{-1/2} \| \\
&= \| L_{\tau} L_{\tau} - \mathcal{D}_{\tau}^{-1/2} D_{\tau}^{1/2} L_{\tau} D_{\tau}^{1/2} \mathcal{D}_{\tau}^{-1} D_{\tau}^{1/2} L_{\tau} D_{\tau}^{1/2} \mathcal{D}_{\tau}^{-1/2} \| \\
&= \| L_{\tau} L_{\tau} - L_{\tau} D_{\tau}^{1/2} \mathcal{D}_{\tau}^{-1} D_{\tau}^{1/2} L_{\tau} D_{\tau}^{1/2} \mathcal{D}_{\tau}^{-1/2} + (I - \mathcal{D}_{\tau}^{-1/2} D_{\tau}^{1/2})L_{\tau} D_{\tau}^{1/2} \mathcal{D}_{\tau}^{-1} D_{\tau}^{1/2} L_{\tau} D_{\tau}^{1/2} \mathcal{D}_{\tau}^{-1/2} \| \\
&\leq \| L_{\tau} (L_{\tau} - D_{\tau}^{1/2} \mathcal{D}_{\tau}^{-1} D_{\tau}^{1/2} L_{\tau} D_{\tau}^{1/2} \mathcal{D}_{\tau}^{-1/2}) \| + a(a+1)^2 \\
&\leq \| D_{\tau}^{1/2} \mathcal{D}_{\tau}^{-1} D_{\tau}^{1/2} L_{\tau} (D_{\tau}^{1/2} \mathcal{D}_{\tau}^{-1/2} - I) - (D_{\tau}^{1/2} \mathcal{D}_{\tau}^{-1} D_{\tau}^{1/2} - I)L_{\tau} \| + a(a+1)^2 \\
&\leq a(a+1) + a + a(a+1)^2.
\end{align*}
 
Consequently, joining the results for the five terms, gives the desired bound. With probability at least
$1-\epsilon$, 
\begin{align*}
 \|\tilde{L} - \tilde{\mathcal{L}} \|
&\leq 2a^3 + 5a^2 + 5a + b \\
&\leq 12a + b \\
&= \{ \varpi^{1/2} + 12 (d + \tau)^{-1/2} \} \{3\log (8N/\epsilon)\}^{1/2} .
\end{align*}
Let $\delta \equiv \varpi^{1/2} + 12 (d + \tau)^{-1/2}$, then the bound becomes
\begin{align*}
\|\tilde{L} - \tilde{\mathcal{L}} \| \leq \delta \{3\log (8N/\epsilon)\}^{1/2} .
\end{align*}

\subsection{Proof of Theorem 2}\label{pfEvBound}
Using Lemma 9 from \cite{mcsherry2001spectral}, let $P_{\tilde{L}}$ be the projection onto the span of the first $K$ left singular eigenvectors of $\tilde{L}$. Then, $P_{\tilde{L}}$ is the optimal rank K approximation to $\tilde{L}$ and 
\begin{align*}
\| P_{\tilde{L}} - \tilde{\mathcal{L}} \|_F^2 \leq 8K \| \tilde{L} - \tilde{\mathcal{L}} \|^2. 
\end{align*}
Next, apply the Davis--Kahan Theorem to $\tilde{\mathcal{L}}$ \citep{davis1970rotation}. Let $W \subset \mathbb{R}$ be an interval and define the distance between $W$ and the spectrum of $\tilde{\mathcal{L}}$ outside of $W$ as 
\begin{align*}
 \Lambda = \min\{ |\lambda - r |; \lambda \text{ eigenvalue of } \tilde{\mathcal{L}}, \lambda \notin W, r \in W \}.
\end{align*}
Choose $W=(\lambda_K/2,\infty)$, where $\lambda_K$ is the $K$th eigenvalue of $\tilde{L}$. Then, $\Lambda = \lambda_K/2$. Let $\omega_K$ be the $K$th largest eigenvalue of $\tilde{\mathcal{L}}$, then under the assumption that $\delta \{3 \log(8N/\epsilon)\}^{1/2}  \leq \lambda_K/2$,
\begin{align*}
 |\lambda_K - \omega_K | \leq \delta \{3 \log(8N/\epsilon)\}^{1/2} \leq \lambda_K/2.
\end{align*}
Hence, $\omega_K \in W$, and $U$ has the same dimension as $\mathcal{U}$. The Davis-Kahan Theorem implies,
\begin{align*}
\| U - \mathcal{U} O \|_F &\leq \frac{2^{1/2} \| P_{\tilde{L}} \tilde{L} - \tilde{\mathcal{L}} \|_F}{\Lambda} \\
&\leq \frac{8^{1/2}\| P_{\tilde{L}} \tilde{L} - \tilde{\mathcal{L}} \|_F}{\lambda_K}\\
&\leq \frac{8 K^{1/2}\| \tilde{L} - \tilde{\mathcal{L}} \|}{\lambda_K}\\
&\leq \frac{8 \delta \{3 K \log(8N/\epsilon)\}^{1/2}}{\lambda_K}
\end{align*}
with probability at least $1 - \epsilon$.

\subsection{Proof of Theorem 3}\label{pfMcRate}
This proof follows the arguments given in \cite{qin2013regularized}.
Let $P= \max_{i} (Z^TZ)_{ii}$ and
\begin{align*}
||\mathcal{C}_i - \mathcal{C}_j ||_2 &\geq ||Z_i (Z^TZ)^{-1/2} V - Z_j (Z^TZ)^{-1/2} V ||_2 \\
&\geq 2^{1/2} || Z^TZ ||_2 \\
&\geq \left(\frac{2}{P}\right)^{1/2} .
\end{align*}
For $\forall Z_j \neq Z_i$, a sufficient condition for one observed centroid to be closest to the population centroid is
\begin{align*}
 ||C_i\mathcal{O}^T-\mathcal{C}_i||_2 < \frac{1}{(2P)^{1/2}} \Rightarrow ||C_i \mathcal{O}^T - \mathcal{C}_i||_2 < ||C_i \mathcal{O}^T-\mathcal{C}_j||_2 ,
\end{align*}
since
\begin{align*}
 ||C_i\mathcal{O}^T-\mathcal{C}_i||_2 < \frac{1}{(2P)^{1/2}} \Rightarrow ||C_i\mathcal{O}^T-\mathcal{C}_j||_2 &\geq || \mathcal{C}_i - \mathcal{C}_j ||_2 - ||C_i\mathcal{O}^T-\mathcal{C}_i||_2 \\
 &\geq \left(\frac{2}{P}\right)^{1/2} - \left(\frac{1}{2P}\right)^{1/2} \geq \frac{1}{(2P)^{1/2}} .
\end{align*}
Let $\mathcal{G}=\{i:||C_i \mathcal{O}^T-\mathcal{C}_i||_2 \geq \frac{1}{(2P)^{1/2}}\}$, so $\mathcal{M} \subset \mathcal{G}$.
Define $Q \in \mathbb{R}^{N\times K}$ where the $i$th row is $C_i$. By the definition of k-means, $||U - Q||_2 \leq ||U - \mathcal{U}\mathcal{O}||_2$. Applying the triangle inequality gives
\begin{align*}
 ||Q - Z\mu\mathcal{O}||_2 = ||Q - \mathcal{UO}||_2 \leq ||U - Q||_2 + ||U - \mathcal{UO}||_2 \leq 2||U -\mathcal{UO}||_2 .
\end{align*}
So,
\begin{align*}
 \frac{||\mathcal{M}||}{N} \leq \frac{||\mathcal{G}||}{N} &= \frac{1}{N} \sum_{i \in \mathcal{G}} 1 \\ 
&\leq \frac{2P}{N} \sum_{i \in \mathcal{G}} ||C_i\mathcal{O}^T - \mathcal{C}_i||^2_2\\
&= \frac{2P}{N} \sum_{i \in \mathcal{G}} ||C_i - Z_i \mu \mathcal{O}||_2^2 \\
&\leq \frac{2P}{N} ||Q-Z \mu \mathcal{O}||_F^2 \\
&\leq \frac{8P}{N} ||U - \mathcal{U}\mathcal{O}||_F^2.
\end{align*}
Thus, using the result from Theorem 2, with probability at least $1-\epsilon$,
\begin{align*}
 \frac{||\mathcal{M}||}{N} \leq \frac{c_0 K P \delta^2 \log (8N/\epsilon)}{N\lambda_k^2},
\end{align*}
where $c_0 = 3 \times 2^6$.

\subsection{Proof of Corollary 1}\label{misCCov}
In order to investigate the mis-clustering bound and the accompanying conditions, we make some simplifying assumptions. Assume $B_{i,i} = p, \forall i$ and $B_{i,j} = q, \forall i \neq j$; in addition, $M_{i,i} = m_1, \forall i$; $M_{i,j} = m_2, \forall i \neq j$; and $R>1$. For computational convenience, assume that each block has the same number of nodes $N/K$ and $R$ is a multiple of $K$. Recall, $\tilde{\mathcal{L}} = Z(\mathcal{D}_B^{-1/2} BZ^T\mathcal{D}_{\tau}^{-1}ZB \mathcal{D}_B^{-1/2} + \alpha M M^T) Z^T  = Z \tilde{B} Z^T$. Therefore, 
\begin{align*}
\tilde{B} =& \left[\frac{1}{N\{p+(K-1)q\}/K + \tau} \right]^2 \left( \frac{N}{K} \right) [(p-q)^2 I + \{2pq + (K-1)q\} \mathbf{1}_K \mathbf{1}_K^T] \\
&+ \alpha \{(m_p-m_q)I + m_q \mathbf{1}_K \mathbf{1}_K^T\},
\end{align*} 
where $m_p = R\{m_1^2+(K-1)m_2^2\}/K$ and $m_q = R\{2m_1m_2+(K-2)m_2^2\}/K$.
For matrices of the form $aI + b\mathbf{1}_K \mathbf{1}_K^T$, $\lambda_K = a$. Note that $m_p - m_q = R(m_1 - m_2)^2/K$. Thus,
\begin{align*}
\lambda_K(\tilde{B}) = \left[ \frac{p-q}{N\{p+(K-1)q\}/K + \tau} \right]^2 \left( \frac{N}{K} \right) + \alpha R(m_1 - m_2)^2/K.
\end{align*}

Recall that $\tilde{\mathcal{L}}$ has the same eigenvalues as $(Z^TZ)^{1/2} \tilde{B} (Z^TZ)^{1/2} = (N/K)^{1/2}I \tilde{B} (N/K)^{1/2}I = (N/K) \tilde{B}$. Hence, the population eigengap is
\begin{align*}
\lambda_K(\tilde{\mathcal{L}}) = \left\{ \frac{p-q}{p+(K-1)q + K \tau/N} \right\}^2 + \frac{\alpha N R (m_1 - m_2)^2}{K^2} .
\end{align*}
Hence, the mis-clustering bound for a growing number of covariates is given by 

\begin{align*}
\frac{|\mathcal{M}|}{N} \leq \frac{\{(d+\tau)^{-1}+\alpha (d+\tau)^{-1/2}\Theta(NR^{1/2})+\alpha^2 \Theta(N^2R)\}\Theta(\log N)}{\alpha \Theta(N^2R^2)+\alpha \Theta(NR)+\Theta(1)}.
\end{align*}

Two conditions in Theorem 3 depend on $R$. Condition $(iii)$ becomes $ R > \Theta(\log N)$ and condition $(iv)$ becomes $\{\alpha N R^{1/2} + (d+\tau)^{-1/2}\}(\log N)^{1/2} \leq \alpha N R + c_0$, which is satisfied for $R \geq \Theta(\log N)$.

Let $R = \Theta\{(\log N)^{a+1}\}$, $d+\tau = \Theta\{(\log N)^{b+1}\}$, and $\alpha = \Theta\{N^{-1}(\log N)^{-1-c}\}$, where $a,b,c \geq 0$, then the mis-clustering rate becomes

\begin{align*}
\frac{|\mathcal{M}|}{N} \leq c_2\frac{(\log N)^{a-2c}+(\log N)^{(a-b)/2-c}+(\log N)^{-b}}{(\log N)^{2(a-c)}+(\log N)^{a-c}+\Theta(1)}.
\end{align*}

If $c$ is chosen such that $a>c$, then $(\log N)^{2(a-c)}$ is the dominant term in the denominator and
$|\mathcal{M}|/N = O\{(\log N)^{-a}\} + O\{(\log N)^{-(3a+b)/2-3c}\} + O\{(\log N)^{2(c-a)-b}\}$.
The mis-clustering rate is minimized when $c=0$, so the rate becomes
$|\mathcal{M}|/N = O\{(\log N)^{-a}\}$.

If $c$ is chosen such that $a \leq c$, 
$|\mathcal{M}|/N = O\{(\log N)^{a-2c}\} + O\{(\log N)^{(a-b)/2-c}\}+ O\{(\log N)^{-b}\}$.
The mis-clustering rate is minimized when $c=\frac{a+b}{2}$, so the rate becomes
$\frac{|\mathcal{M}|}{N} = O\{(\log N)^{-b}\}$.

Hence, to minimize the mis-clustering rate when $a\leq b$ choose $c=\frac{a+b}{2}$, which yields a mis-clustering rate of $O\{(\log N)^{-b}\}$, and when $a>b$ choose $c=0$, which gives a mis-clustering rate of $O\{(\log N)^{-a}\}$. If we consider the special case where $a = 0$ or $R = \Theta(\log N)$ and $b = 0$ or $d + \tau = \Theta(\log N)$. The theoretical results above suggest $\alpha = \Theta\{(N \log N)^{-1}\}$. This result agrees with the value suggested by the empirical procedure in \S 2$\cdot$3, which yields $\alpha_{\min} = \alpha_{\max} = \Theta\{(N \log N)^{-1}\}$ when $R=\Theta(\log N)$ based on the population eigenvalues.

\subsection{Proof of Corollary 2}\label{cor2}
Perfect clustering requires that $|\mathcal{M}| < 1$. Based on the bound in Theorem 3, this corresponds to $\delta \{c_0KP\log(8N/\epsilon)\}^{1/2} < \lambda_K$. Under the same simplifying assumptions as above, this becomes
\begin{align*}
c' \{\alpha N R^{1/2} + (d + \tau)^{-1/2}\} (N\log N)^{1/2} &< \alpha N R + \Theta(1), \\
c'' \alpha N R^{1/2} (N\log N)^{1/2} &< \alpha N R, \\
R &\geq \Theta(N \log N).
\end{align*}

\subsection{Proof of Theorem 4}\label{pfLowerBound}
This proof uses Fano's inequality to derive the lower bound following an approach similar to \cite{chaudhuri2012spectral}. Let $G_S$ be a partition given by a specific $S$, the set of all nodes in the first block, and let $F$ be the family of all such partitions. Fano's inequality states
\begin{align*}
 \sup_{G_S \in F} P_{G_S} (\Psi \neq G_S) \geq 1 - \frac{\beta + \log 2}{\log r},
\end{align*}
where $KL(G_S, G_{S'}) \leq \beta$, $r = |F| - 1$, and $\Psi$ is the estimated node partition based on the observed edges and node covariates. 

First, by independence the KL-divergence can be written as follows,
\begin{align*}
KL(G_S, G_{S'}) = \sum_{e \in E} KL(\rho_e, \rho_e') + \sum_{v \in V} KL(\gamma_v, \gamma_v').
\end{align*}
Let $\rho_e$ and $\rho_e'$ be the distribution for edge $e$ and $\gamma_v$ and $\gamma_v'$ be the covariate distribution for node $v$ in $G_{S}$ and $G_{S'}$, respectively. Recall $B_{1,1} \geq B_{2,2} \geq B_{1,2}$ and let $b_i \in \{ B_{1,1}, B_{2,2}, B_{1,2} \}$.
For a single edge when $\rho_e \neq \rho_e'$, 
\begin{align*}
 KL(\rho_e, \rho_e') &\in \{ b_i \log \frac{b_i}{b_j} + (1-b_i) \log \frac{1-b_i}{1-b_j} \}\\ 
 &\leq  B_{1,1} \log \frac{B_{1,1}}{B_{1,2}} + (1-B_{1,1}) \log \frac{1-B_{1,1}}{1-B_{1,2}} + 
		      B_{1,2} \log \frac{B_{1,2}}{B_{1,1}} + (1-B_{1,2}) \log \frac{1-B_{1,2}}{1-B_{1,1}}\\
&=  (B_{1,1} - B_{1,2}) \log \left\{1 + \frac{B_{1,1}-B_{1,2}}{B_{1,2}(1-B_{1,1})} \right\}\\ 
&\leq \frac{(B_{1,1} -B_{1,2})^2}{B_{1,2}(1-B_{1,1})}.
\end{align*}
 Now find the KL-divergence of the covariates on a single node. For $\gamma_v \neq \gamma_v'$,
\begin{align*}
 KL(\gamma_v, \gamma_v') &= \sum_j^R KL(\gamma_{v_j}, \gamma'_{v_j}) \equiv \Gamma .
\end{align*}
For the case of Bernoulli random variables where the $j$th covariate has probability $M_{1,j}$ in block 1 and $M_{2,j}$ in block 2, this is
\begin{align*}
 KL(\gamma_{v_j}, \gamma_{v_j}') &= 
\begin{cases}
M_{1,j} \log \frac{M_{1,j}}{M_{2,j}} + (1-M_{1,j}) \log \frac{1-M_{1,j}}{1-M_{2,j}} \text{, } & v \in \text{block 1}\\
M_{2,j} \log \frac{M_{2,j}}{M_{1,j}} + (1-M_{2,j}) \log \frac{1-M_{2,j}}{1-M_{1,j}} \text{, } & v' \in \text{block 1} \\
\end{cases} \\
&\leq (M_{1,j}-M_{2,j})\log \frac{M_{1,j}(1-M_{2,j})}{M_{2,j}(1-M_{1,j})}.
\end{align*}
Therefore, the KL-divergence is bounded by
\begin{align*}
 KL(G_S, G_{S'}) &\leq \binom{N}{2} \frac{(B_{1,1} -B_{1,2})^2}{B_{1,2}(1-B_{1,1})} + N\Gamma 
\leq \frac{N^2}{2} \frac{(B_{1,1} -B_{1,2})^2}{B_{1,2}(1-B_{1,1})} + N\Gamma . 
\end{align*}
The number of partitions can be bounded as follows,
\begin{align*}
|F|&= \frac{1}{2} \binom{N}{N/2} = \frac{N!}{2\{(N/2)!\}^2} \\
&\geq \frac{(2\pi N )^{1/2}(N/e)^{N}}{[e(N/2)^{1/2}\{N /(2e)\}^{N/2}]^2}\\
&\geq \frac{2^{N-2.1}}{(N/2)^{1/2}},
\end{align*} 
where the first inequality uses $(2\pi N)^{1/2}(N/e)^N \leq N! \leq e N^{1/2} (N/e)^N$.
Now the $\log$ term is bounded by
\begin{align*}
\log (|F|-1) &\geq \log \left\lbrace \frac{2^{N-2.1}}{(N/2)^{1/2}} - 1\right\rbrace \\
&\geq (N-3)\log 2 - \frac{1}{2} \log(N/2) \\
&\geq \frac{\log 2}{2} N \text{ for } N \geq 8 .
\end{align*}
Thus, by Fano's inequality, in order to correctly determine the block assignments with probability at least $1-\epsilon$ requires
\begin{align*}
 \epsilon &\geq 1 - \frac{ N^2 (B_{1,1} - B_{1,2})^2/\{2B_{1,2}^2(1-B_{1,1})^2\} +N\Gamma + \log 2}{(N\log 2)/2}, \\
B_{1,1} -B_{1,2} & \geq B_{1,2}(1-B_{1,1}) [\frac{2}{N} \left\{\frac{\log 2}{2}(1-\epsilon) - \Gamma - \frac{\log 2}{N} \right\}]^{1/2} .
\end{align*}
Fix $B_{1,1}$ and let $\Delta = B_{1,1} - B_{1,2}$, then rewrite this bound as 
\begin{align*}
 \Delta \geq \frac{B_{1,1}(1-B_{1,1})}{\left[\frac{2}{N} \left\{\frac{\log 2}{2}(1-\epsilon) - \Gamma - \frac{\log 2}{N} \right\}\right]^{-1/2}+(1-B_{1,1})}.
\end{align*}

\subsection{Comparison of the General Lower Bound to Theorem 3} \label{boundComp}
First, simplify the general lower bound given in Theorem 4 to make the comparison with Theorem 3 easier.
\begin{align*}
\Delta &\geq \frac{B_{1,1}(1-B_{1,1})}{\left[\frac{2}{N} \left\{\frac{\log 2}{2}(1-\epsilon) - \mathcal{K} - \frac{\log 2}{N} \right\}\right]^{-1/2}+(1-B_{1,1})} \\
&\geq \frac{B_{1,1}(1-B_{1,1})}{3/2\left[\frac{2}{N} \left\{\frac{\log 2}{2}(1-\epsilon) - \mathcal{K} - \frac{\log 2}{N} \right\}\right]^{-1/2}} \\
&\geq B_{1,1}(1-B_{1,1})\left(\frac{2}{3}\right)\left[\frac{2}{N} \left\{\frac{\log 2}{2}(1-\epsilon) - \mathcal{K} - \frac{\log 2}{8} \right\}\right]^{1/2} \\
&\geq \frac{c_4}{N^{1/2}}.
\end{align*}
According to Theorem 3 to achieve perfect clustering with probability $1 - \epsilon$, requires $\delta \{c_0 K P  \log(8N/\epsilon)\}^{1/2} < \lambda_K$. As shown in \S \ref{cor2}, this requires $R \geq \Theta(N \log N)$.

\bibliographystyle{biometrika}
\bibliography{casc}

\end{document}